\documentclass[10pt,twocolumn,letterpaper]{article}

\usepackage[pagenumbers]{cvpr} % To force page numbers, e.g. for an arXiv version

\usepackage{algorithm}
\usepackage{algorithmic}
\usepackage{multirow}
\usepackage[table]{xcolor}  
\usepackage{threeparttable}
\usepackage{adjustbox}

\definecolor{cvprblue}{rgb}{0.21,0.49,0.74}
\usepackage[pagebackref,breaklinks,colorlinks,allcolors=cvprblue]{hyperref}

\def\paperID{39302} % *** Enter the Paper ID here
\def\confName{CVPR}
\def\confYear{2026}

\title{Q-MambaIR: Accurate Quantized Mamba for Efficient Image Restoration}

\author{
Yujie Chen\textsuperscript{1},
Haotong Qin\textsuperscript{1}\dag,
Zhang Zhang\textsuperscript{2},
Michele Magno\textsuperscript{1},
Luca Benini\textsuperscript{1},
Yawei Li\textsuperscript{1,3}\dag\\[0.5em]
\textsuperscript{1}ETH Z\"urich \quad
\textsuperscript{2}Shenzhen Automotive Research Institute \quad \textsuperscript{3}Nanyang Technological University\\[0.3em]
{\tt\small \{li.yawei.ai@gmail.com, haotong.qin@pbl.ee.ethz.ch\}}
}
\begin{document}

\twocolumn[{%
\renewcommand\twocolumn[1][]{#1}%
\maketitle
\begin{center}
    \includegraphics[width=.9\linewidth]{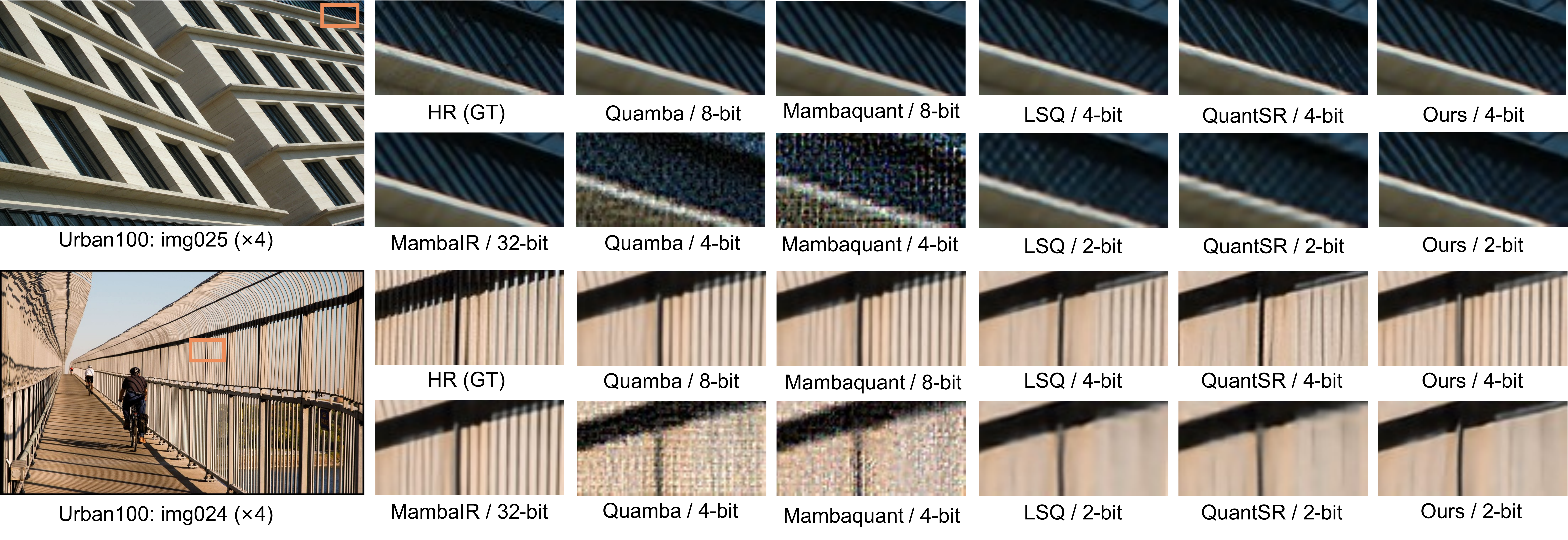}%
    \vspace{-2.3mm}
    \captionof{figure}{Visual comparison ($4 \times$ upscaling) with quantized MambaIR in terms of 4-bit and 2-bit quantization. We compare our Q-MambaIR with recent quantization methods (\textit{i.e}., Quamba \cite{chiang2025quamba}, MambaQuant \cite{xu2025mambaquant}, LSQ \cite{esser2019learned}, QuantSR \cite{qin2023quantsr}). Our Q-MambaIR exhibits clear and consistent advantages over all compared methods.} %\yawli{Put this figure here. Use pdf formats. Select two good-looking example images here.}
    \label{fig:teaser}
\end{center}
}]
\begin{abstract}
    State-space models (SSMs) have attracted considerable attention for image restoration (IR) tasks due to their ability to scale linearly with sequence length while effectively capturing long-distance dependencies. However, deploying SSMs to edge devices is challenging due to constraints in memory, computing capacity, and power consumption, underscoring the need for efficient compression strategies. While low-bit quantization is an efficient and hardware-friendly model compression strategy for reducing model size and accelerating IR tasks, SSMs suffer from substantial performance degradation at ultra-low bitwidths (2--4 bits), primarily due to outliers that exacerbate quantization error. To address this challenge, we propose {\bf Q-MambaIR}, an accurate, efficient, and flexible {\bf Q}uantized {\bf Mamba} for {\bf IR} tasks. Specifically, we introduce a statistical \textit{dynamic-balancing learnable scalar} (DLS) to dynamically adjust the mapping range for activation quantization, thereby mitigating the peak truncation loss caused by extreme values. Furthermore, we design a \textit{range-floating flexible allocator} (RFA) with an adaptive threshold to round the values flexibly. This approach preserves high-frequency details while maintaining the SSM’s feature extraction capability. Notably, RFA also enables pre-deployment weight quantization, striking a balance between computational efficiency and model accuracy. Extensive experiments on IR tasks demonstrate that Q-MambaIR consistently outperforms existing quantized SSMs, establishing the new state-of-the-art (SOTA) accuracy with only negligible additional training computation and storage requirements. Code will be released at https://github.com/Areache/Q-MambaIR.
\end{abstract}    
\section{Introduction}
\label{sec:intro}
Image restoration (IR) has been widely recognized as an important problem in computer vision and image processing~\cite{gu2015convolutional,gu2014weighted,kim2016deeply, lim2017enhanced,zamir2021multi,liang2021swinir,zamir2022restormer,guo2024mambair}. It focuses on recovering 
high-quality images from low-quality versions corrupted by noise, blur, or compression artifacts. The state-of-the-art methods are based on deep neural networks (DNNs), combining convolutional neural networks (CNNs)~\cite{lim2017enhanced,ledig2017photo} for local feature extraction and transformers~\cite{liang2021swinir,li2023efficient, zhang2024transcending} for modeling long-range dependencies within images. Recently, state space models (SSMs)~\cite{gu2023mamba,guo2024mambair} have been proposed to model long-range dependencies with linearly scaled computation relative to the sequence length. 
% SSM obtains the computationally friendly global receptive field compared to CNNs and transformers, through . 
However, the huge model size of the SSM-based architecture remains a challenge for real-time deployment on resource-constrained devices. This calls for effective and principled SSM compression approaches to reduce computational costs, which remain largely underexplored in the IR field.

% \begin{figure}
%   \centering
%   % \fbox{\rule{0pt}{2in} \rule{.9\linewidth}{0pt}}
%   \includegraphics[width =.5\linewidth]{leverage.pdf}
%   \caption{Q-MambaIR significantly improves the efficiency of state sapce models for IR.}
%   \label{fig:balance}
% \end{figure}

\begin{figure}
  \centering
    \includegraphics[width =.9\linewidth]{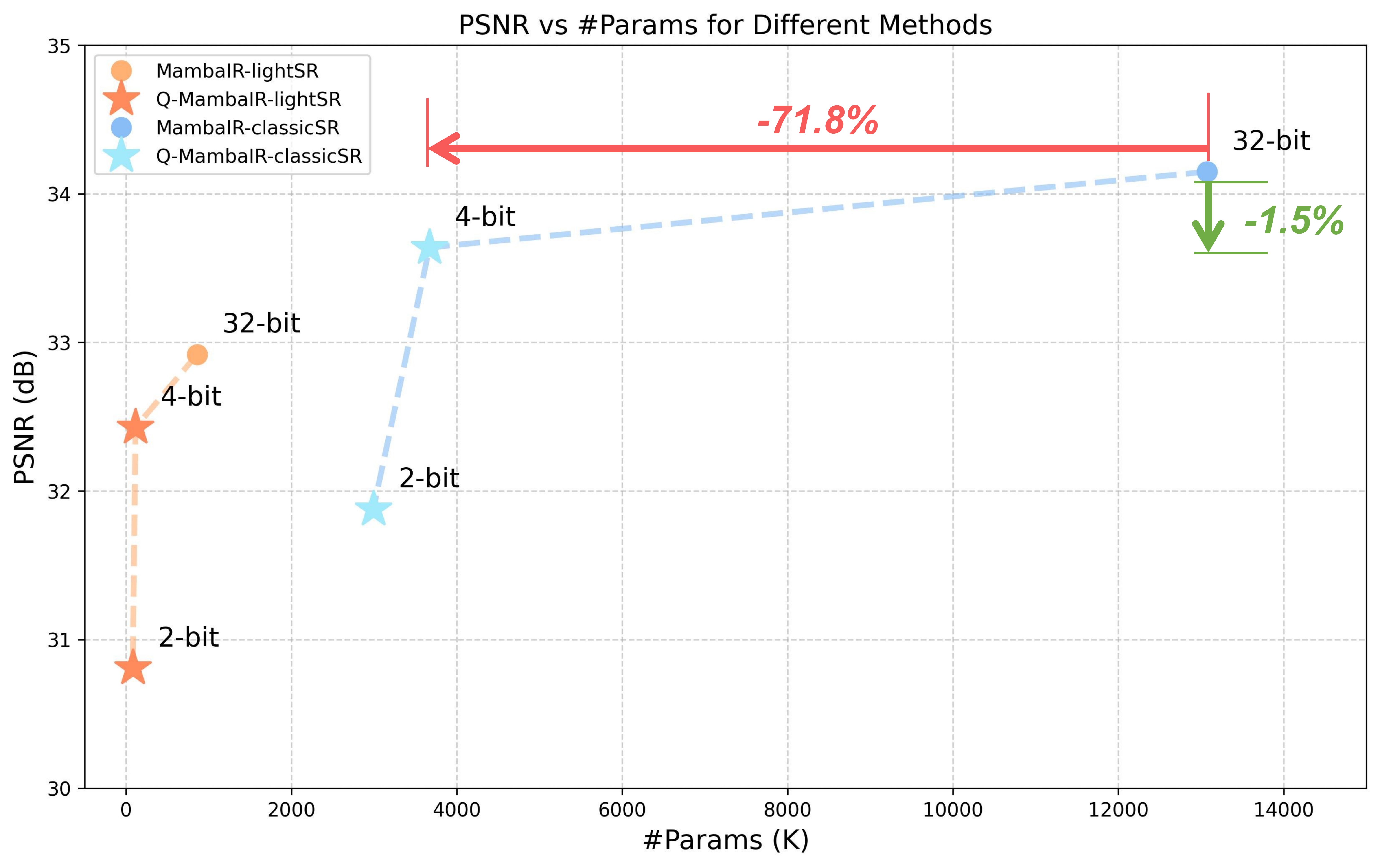}
  % \vspace{-4mm}
    \caption{Q-MambaIR significantly improves the efficiency of state space models for image restoration. Results are shown for 2$\times$ super-resolution over the Urban100 benchmark: the orange curve corresponds to lightweight SR and the blue curve to classic SR.}
  \vspace{-4mm}
  \label{fig:balance}
  % \vspace{-2mm}
\end{figure}

Low-bit quantization has emerged as an effective approach to compress DNNs and accelerate their inference~\cite{li2021brecq,ding2022towards,qin2023quantsr}. By converting model weights and activations from high-precision floating-point formats (such as FP32) to lower-bitwidth integer representations, typically 8-bit, 4-bit, or even fewer, the overall memory footprint and computational cost can be significantly reduced. 
Post-training quantization (PTQ) applies quantization to a pre-trained model without retraining, offering rapid deployment but exhibiting notable performance degradation at ultra-low bitwidths~\cite{xu2025mambaquant,chiang2025quamba}. In contrast, quantization-aware training (QAT) incorporates quantization effects during training, allowing the model to adapt to quantization-induced perturbations and thereby achieve higher accuracy under comparable bitwidth constraints~\cite{choi2018pact,esser2019learned,wang2023bitnet}.

\begin{figure}
  \centering
    % \fbox{\rule{0pt}{2in} \rule{.9\linewidth}{0pt}}
    \includegraphics[width =\linewidth]{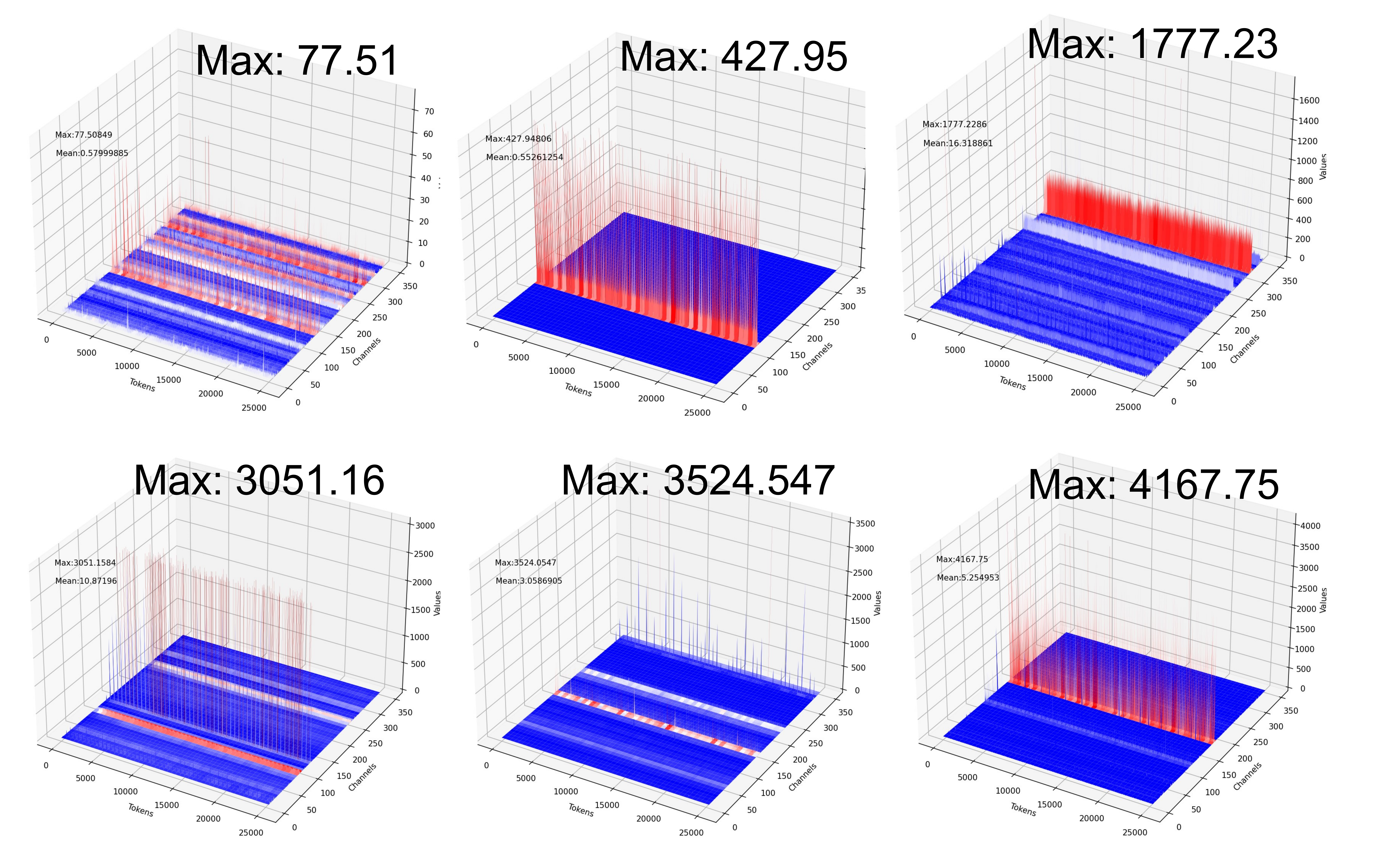}
  % \vspace{-4mm}
    \caption{Outlier visualization in SS2D. The outliers exist in output
 from the selective SSM. With the quantization step, significant errors are easily caused in the output SSMs after the linear recurrent system \cite{chiang2025quamba}. Severe performance degradation occurs in SSMs quantization.}
  % \vspace{-4mm}
  \label{fig:outlier}
  \vspace{-3mm}
\end{figure}

Despite the potential, ultra-low bitwidths quantization of SSMs presents a significant challenge, resulting in severe quantization failures and performance degradation.
\textbf{First}, inherent structural outliers occur in SSMs (see Fig.~\ref{fig:outlier}). Combined with dynamically varying inputs, they pose a fundamental dilemma for conventional static quantization: expanding the quantization range to include these outliers reduces the resolution in dense-value regions, while narrowing it risks truncating informative outliers. This trade-off remains unresolved in a fixed quantization range.
 % \yawli{There should be one figure that explains this phenomenon.}
\textbf{Second}, commonly used deterministic rounding in low-bit quantization introduces forward parameter homogenization and backward gradient mismatch, leading to the accumulation of discretization errors, which degrade feature representations. Such limitations hinder the model’s ability to maintain a balance between precision and computational efficiency. The balance is important in image restoration tasks, where both fine-grained textures and sparse, extreme-value activations play a critical role in performance.

Inspired by the above insights, we propose an accurate, efficient, and flexible \textbf{Q}uantized \textbf{Mamba} for \textbf{I}mage \textbf{R}estoration (\textbf{Q-MambaIR}). We first propose a \textit{Dynamic-balancing Learnable Scalar} (DLS), where we design a learnable scalar to dynamically adapt the quantization range by capturing activation distribution characteristics. This mechanism enables a dynamic trade-off between preserving fine-grained details and retaining informative outliers. We further propose a \textit{Range-floating Flexible Allocator} (RFA), which is a flexible discretization mechanism with a flexible allocator using adaptive thresholds to round the values. RFA proves to be effective in mitigating parameter homogenization and error accumulation through the preservation of local gradients, thus maintaining the feature representation capacity of the SSM block.

Extensive experiments demonstrate that Q-MambaIR consistently outperforms advanced quantization methods across various bitwidths (Fig.~\ref{fig:balance}). In particular, we show that Q-MambaIR achieves superior performance compared to existing quantization methods tailored for SSM, validating its effectiveness in addressing the quantization challenges inherent to SSM architectures. Our main contributions are summarized as follows.

\begin{itemize}

% \item We explore the theocratically reason of SSM quantization degradation and experimentally finding the challenges in quantifying SSM.\\
\item We propose Q-MambaIR, an accurate quantized Mamba for efficient image restoration. It narrows the gap between low-bit quantized models and their full-precision counterparts, with the potential to facilitate deployment in resource-constrained scenarios.

\item To preserve the extreme value, we propose a DLS. Specifically, our DLS effectively reduces the loss of truncation by capturing the statistical characteristics of the activation and dynamically adjusting the quantization ranges.
% \item From activation in SSM, we find that the reason for the quantization loss of ss2d is mainly in the output activation quantization, and thus we propose a low-loss quantizer for QAT with high flexibility and high adaptability.

\item To enhance high-frequency feature retention, we propose an RFA, which preserves critical details through adaptive threshold-based discretization to enhance feature extraction capabilities, thereby improving restoration quality.

\item We employ our Q-MambaIR to compress SSM-based IR networks under low-bit settings while maintaining competitive restoration quality, resulting in the corresponding quantized baseline. Our Q-MambaIR achieves superior performance over SOTA methods.

\end{itemize}

\section{Method}
\label{sec:method}
\begin{figure*}
  \centering
    \includegraphics[width =.9\linewidth]{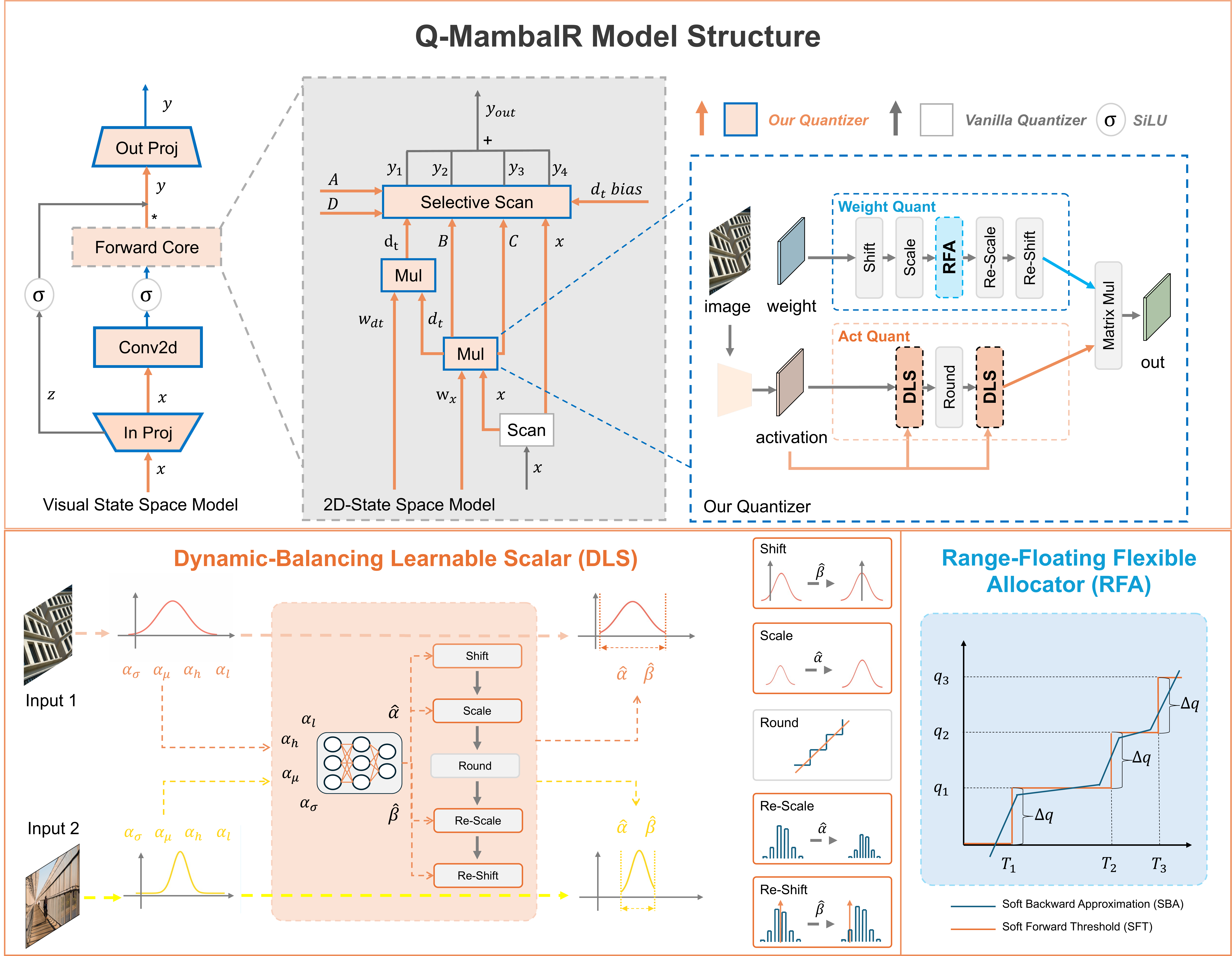}
    \caption{\textbf{Upper Part:} Overview of the proposed Q-MambaIR structure. It consists of three hierarchical levels: (1) the vision state space model (VSSM) as the overall architecture, (2) the 2D state space model (2D-SSM) as its core computational module, and (3) our custom quantization operators. We utilize RFA-based flexible rounding for weights and DLS-based shifting and scaling for activations, integrated into the operators of the VSSM block. \textbf{Lower Part:} Operator-level quantization design in Q-MambaIR. The left part illustrates the Dynamic-balancing Learnable Scalar (DLS) for activation quantization, which balances the quantization range to trade off between preserving outliers in wide-range inputs and retaining details in narrow-range inputs. The right part shows the Range-floating Flexible Allocator (RFA) for weight quantization. RFA performs a transformation from the input range, guided by learnable thresholds, to a uniform quantized interval, while approximating backpropagation through adaptive, high-similarity piecewise functions.}
    % \vspace{-3mm}
    \label{fig:framework}
  \label{fig:qssm}
  \vspace{-2mm}
\end{figure*}

To provide a solution for SSM low-bit quantization, we first introduce an overview of the main part of VSSM employed in MambaIR, followed by an analysis of the limitations inherent in existing low-bit quantization architectures. Based on this analysis, we propose our \textbf{Q-MambaIR} to address these limitations. Fig.~\ref{fig:qssm} shows an overview of the proposed quantization framework, illustrating the network structure, the design of the quantization operator, and detailed quantization flows for both activations and weights. For activation quantization, DLS overcomes the constraints of deterministic quantization ranges by dynamically adapting to the distribution of activations, thereby minimizing the truncation loss caused by outlier clipping. For weight quantization, RFA introduces a soft rounding mechanism with a flexible rounding threshold, which mitigates the accumulated errors typically associated with static rounding schemes.

\subsection{Preliminaries}

\noindent \textbf{Visual State Space Model.} The VSSM module is the main functioning feature extractor in visual Mamba for IR. The architecture of VSSM is demonstrated in Fig.~\ref{fig:qssm}. It adapts SSM to vision data without compromising its advantages.
Specifically, SSM defines a sequence-to-sequence transformation $x(t) \in \mathbb{R} \rightarrow y(t) \in \mathbb{R}$ through a hidden state $h(t) \in \mathbb{R}^{N}$. It has two stages and captures long-range dependencies with the state-space equation, namely,
\begin{align}
    h(t) &= A h(t-1) + B x(t), \\
        y(t) &= C h(t) + D x(t),
    \label{eq:ssm}
\end{align}
where $N$ is the state size, $A \in \mathbb{R}^{N \times N}$, $B \in \mathbb{R}^{N \times 1}$,  $C \in \mathbb{R}^{1 \times N}$, and $D \in \mathbb{R}$. 
For efficient, parallelizable computation during training, the module can be formulated in a convolutional form, namely,
\begin{align}
    \label{eq:cnn-form}
    \overline{K} &\triangleq \left( C \overline{B}, C \overline{A} \overline{B}, \dots, C \overline{A}^{L-1} \overline{B} \right), \\
    y &= x \circledast \overline{K}.
\end{align}
% \subsubsection{Quantization Framework}
\noindent \textbf{Quantization of VSSM.} As a common practice in quantization, the weights and activations used in the computation units are mapped to a uniform discrete set, namely,
\begin{equation}
        Q(\textbf{x}, \alpha, \beta) = \frac{\lfloor
        \operatorname{clip} \left ((x - \beta) \cdot \alpha \right)
        \rfloor}
        {\alpha}+\beta,
    \end{equation}
where $Q(\textbf{x}, \alpha, \beta)$ represents the quantized weights or activations with low bitwidth, and $x$ is the input tensor in floating-point format. The function $\operatorname{clip}(x) = \max(\min(x, 2^{n-1} - 1), -2^{n-1})$ ensures that the values are limited to the representable range of $n$-bit integers before rounding.  $\alpha$ is the scaling factor that determines how the range of $X$ is mapped to the rounding range between $-2^{n-1}$ and $2^{n-1} - 1$. $\beta$ is the shifting factor that adjusts the zero-point alignment, affecting the distribution of the quantization range.

% Since the rounding function results in all-zero gradients and blocks backpropagation, straight-through estimation (STE)[??] is used to approximate the gradients of discrete variables during training.

% \begin{figure*}
%   \centering
%     \includegraphics[width =.9\linewidth]{contribution.pdf}
%     \caption{Operator-level quantization design in Q-MambaIR. The left part illustrates the Dynamic-balancing Learnable Scalar (DLS) for activation quantization, which balances the quantization range to trade off between preserving outliers in wide-range inputs and retaining details in narrow-range inputs. The right part shows the Range-floating Flexible Allocator (RFA) for weight quantization. RFA performs a transformation from the input range, guided by learnable thresholds, to a uniform quantized interval, while approximating backpropagation through adaptive, high-similarity piecewise functions.}
%     % \vspace{-3mm}
%   \label{fig:detail}
%   % \vspace{-2mm}
% \end{figure*}

\subsection{Dynamic-Balancing Learnable Scalar}

\noindent\textbf{Why input-adaptive quantization for SSMs?}
Unlike CNNs, where activation distributions are relatively stable across inputs, SSMs compute input-dependent projections $\mathbf{B}(\mathbf{x}){=}W_B\mathbf{x}$ and $\Delta(\mathbf{x}){=}\mathrm{softplus}(W_\Delta\mathbf{x})$, causing activation statistics to vary significantly per input. This poses a fundamental challenge for any quantizer with a static or learned-fixed range: classical quantization theory~\cite{banner2019post,nagel2021whitepaperneuralnetwork} shows that the expected MSE of a $b$-bit uniform quantizer is minimized when its range $[\alpha,\beta]$ matches the input distribution. When activation statistics $(\mu_x, \sigma_x)$ vary across inputs, a fixed range incurs an excess error proportional to $\mathrm{Var}(\mu_x){+}\mathrm{Var}(\sigma_x)$, while an input-adaptive predictor reduces this excess to a small prediction residual. Since SSMs' input-dependent matrices induce large activation variance across inputs, the adaptive approach yields substantially lower quantization error than any fixed-range method. This analysis motivates the design of DLS. A complete derivation is provided in the \textit{Supplementary Material}.

\noindent\textbf{DLS formulation.}
Building on the above analysis, DLS predicts the quantization range $(\hat{\alpha}, \hat{\beta})$ from input statistics rather than learning them as fixed parameters. We extract four statistical descriptors from each input tensor and feed them through a lightweight linear predictor, establishing an explicit mapping between the optimal quantization range and the inherent distribution of the data. Formally, the DLS quantizer is defined as:
\begin{equation}
    \begin{aligned}    
    Q_\text{DLS}(\mathbf{x}, \hat{\alpha},\hat{\beta}) &= \frac{\lfloor
    \operatorname{clip} \left ((x - \hat{\beta}) \cdot \hat{\alpha} \right)
    \rfloor}{\hat{\alpha}}+\hat{\beta},\\ 
    \text{with} \quad \hat{\alpha} &= \left| \mathbf{w}_1^\top \Phi'(x) \right|, \quad
        \hat{\beta} = \mathbf{w}_2^\top \Phi(x),
    \end{aligned}
\end{equation}
where the composite feature vector $\Phi(x) = (\mu_x, \sigma_x, x_{\min}, x_{\max})$ captures four statistical descriptors of the input, and its variant $\Phi'(x) = (|\mu_x|, \sigma_x, x_{\min}, x_{\max})$ ensures a non-negative scale. The learnable weight vectors $\mathbf{w}_1$ and $\mathbf{w}_2$ are jointly optimized with the network, projecting the statistics into the scaling factor $\hat{\alpha}$ and the shifting factor $\hat{\beta}$ that govern the quantization range. The $Q_{\text{DLS}}$ quantizer is embedded into all activation paths of the quantized VSSM block, as illustrated in Fig.~\ref{fig:framework} (lower-left). The effect of different initialization strategies for $\hat{\alpha}$ and $\hat{\beta}$ is analyzed in \cref{tab:INIT}.

\noindent\textbf{Discussion.}
By tying quantization parameters to input statistics, DLS resolves the fundamental incompatibility between fixed quantization ranges and the dynamic activation distributions of SSMs. When the input contains extreme outliers, the predicted range expands to preserve them; under low-variance conditions, the range contracts to retain a high-resolution representation in dense regions. This adaptive behavior simultaneously reduces outlier-induced clipping loss and improves bit allocation efficiency across diverse inputs.

\subsection{Range-Floating Flexible Allocator}
\label{subsec:rfa}
% 针对SSM参数动态范围大、分布非均匀的特性.存在特征提取困难的特点. 而图像重建任务注重细节丰富,局部特性还原重要.而传统固定round对于局部结构不友好,局部细节特征损失严重.本文提出软舍入量化（Soft Rounding Quantization, SRQ）方法，用于替换传统量化中的rounding,它通过门限自适应机制与梯度近似技术的协同设计，提升量化后weight对数值做灵活的表达的能力,显著提升量化模型精度。
To address the discretization and gradient mismatch bottleneck, RFA employs a threshold-adaptive mechanism that dynamically adjusts the rounding boundary based on tensor statistics. It is combined with a gradient approximation technique to align forward and backward propagation (see \cref{fig:framework} lower part right).
% 一个range-floating 的更加灵活的舍入可以减少离散化误差和向前向后传播不匹配.对于保证ssm中的权重对高频纹理的提取能力,减少细节特征的丢失.增强网络了图像重构的灵活性尤其重要。.并且由于对SSM的激活使用RFA支持部署前量化,能有效提高精度和保证推理效率
% State Space Models (SSMs) typically exhibit large dynamic ranges and non-uniform distributions of parameters, posing significant challenges for traditional quantization methods in feature extraction. Particularly in image reconstruction tasks, where detail richness and local feature restoration are critical, conventional fixed rounding methods are ill-suited for local structures, often leading to severe loss of detail features. It significantly enhancing the flexible representation capability of quantized weights and improving model accuracy.\\
% soft forward threshold:输入舍入function的缩放后数据$x$被映射到最近的量化级$q_i$

% \yawli{CUDA implementation can significantly improve the solidness of the paper.}

% \yawli{The implementation should not be very difficult. You can refer to \url{https://github.com/ShuhangGu/MTLU_ICCV2019}, a lot of implementation detailed can be copied directly from there.}

\noindent\textbf{Soft Forward Threshold} maps the input $w$ to the nearest quantization level $q_i$:
\begin{equation}
   \hat{w}_{\text{forward}} = \sum_{i=1}^{N} q_i \cdot \mathbb{I}\left(w \in [T_i, T_{i+1})\right),
\end{equation}
% \yawli{More details should be revealed.} 
where learnable quantization threshold parameters $ T = \{T_1, T_2, \dots, T_N\}, T_i \in \mathbb{R}$ are jointly optimized with network training, enabling quantization intervals to float adaptively and retain fine-grained gradient information. This is a crucial property for SSMs, which are particularly sensitive to discretization and accumulation errors. 
% \yawli{Additional analysis would be interesting: 1) what's the distribution of $q_i$ after training? 2) what's the distribution of $T_i$ after training? 3) Is this another version of AdaRound?}

\noindent\textbf{Soft Backward Approximation} constructs a differentiable gradient path via piecewise linear interpolation:
\begin{equation}
\frac{\partial \hat{w}}{\partial w} = 
\begin{cases}
    0.1, & \text{if } w \in \mathcal{F} \quad \text{(flat region)} \\
    \dfrac{\Delta q_i}{T_{i+1} - T_i}, &otherwise \quad \text{(transition region)}
\end{cases}
\end{equation}
In backward propagation, the fixed slope (0.1) is applied to approximate regions where the gradient of the round function is zero, preventing the gradient from vanishing. The adaptive slope $\frac{\Delta q_i}{T_{i+1} - T_i}$ is adopted near the quantization transition region (where the gradients approach infinity) to precisely align with discrete quantization levels. Here, $\Delta q_i = q_{i+1} - q_i$ represents the size of the quantization step, constructing a uniformly distributed integer interval.

% 高密度区域分辨率不足：大量激活值集中在狭窄区间时，固定间隔量化级无法充分捕捉细节；
% 低密度区域资源浪费：稀疏区域的量化级未被有效利用，增加量化误差。
% 离散量化级集合：
% \begin{figure*}
%   \centering
%   \includegraphics[width =\linewidth]{ICCV2025-Author-Kit-Feb/visual_comparision_2.png}
%     \caption{Ours method performs in visual comparision in 4-bit and 2-bit quantization. Our Q-MambaIR performs obviously better than others in both 4-bit and 2-bit cases.
% }
% \label{fig:compar2}
% \end{figure*}
% As shown in Figure X (cite), the comparative analysis between RFA and conventional Gradient Rounding (GR) demonstrates the superior performance of RFA in detail preservation and feature extraction: 
% e.g. In image reconstruction tasks, SRQ preserves finer texture details (e.g., edges and patterns) in activation maps, while GR causes high-frequency detail blurring.
The range-floating mechanism reduces discretization errors by flexibly adjusting quantization thresholds in the SSM weights, thus mitigating detail loss during texture extraction. Importantly, the quantized weights strictly adhere to uniform intervals, ensuring direct compatibility with standard fixed-point arithmetic units, thus introducing close to zero overhead during inference. In summary, this co-design of adaptive rounding and gradient optimization enables low-loss compression of high-frequency textures while maintaining inference efficiency. The algorithm details of RFA are in the \textit{Supplementary Material}.
\section{Experiments}

\subsection{Settings}
\subsubsection{Dataset and Evaluation
}
 Following the setup in previous work \cite{guo2024mambair}, we conduct experiments on various IR tasks, including image SR (classic SR and lightweight SR), JPEG compression artifact reduction, and image denoising (i.e., Gaussian color image denoising). For classic SR training, the DIV2K \cite{timofte2017ntire} and Flickr2K \cite{lim2017enhanced} datasets are combined, while lightweight SR models are trained exclusively on DIV2K \cite{timofte2017ntire}. Benchmarking for SR methods employs five standard datasets: Set5 \cite{BMVC.26.135}, Set14 \cite{zeyde2012single}, B100 \cite{martin2001database}, Urban100 \cite{huang2015single}, and Manga109 \cite{matsui2017sketch}. Gaussian denoising models are trained using DIV2K \cite{timofte2017ntire}, Flickr2K \cite{lim2017enhanced}, BSD500 \cite{arbelaez2010contour}, and WED \cite{ma2016waterloo}, with evaluations performed on BSD68 \cite{martin2001database}, Kodak24, McMaster \cite{zhang2011color} and Urban100 \cite{huang2015single}. 
 % Real-world denoising experiments leverage 320 high-quality images from the SIDD \cite{abdelhamed2018high} training dataset, validated on the SIDD test dataset and DND \cite{plotz2017benchmarking} dataset. 
 Quantitative analysis employs PSNR and SSIM metrics computed on the luminance (Y) component of the YCbCr space. We also use a perceptual metric: LPIPS \cite{ghildyal2022shift} to capture perceptual quality. 

\begin{table*}
    \caption{Quantitative comparison on \textbf{classic image SR} tasks with state-of-the-art approaches. Methods annotated with '*' denote PTQ techniques. The best results are highlighted in red, while the second-best results are marked in blue. Additionally, full-precision models are distinguished using gray row color for clarity.}
    % \vspace{-2mm}
  \label{tab:classic}
    \small
    \centering
  \begin{adjustbox}{width=\textwidth}
    \begin{tabular}{@{}l c c c c c c c c c c c c c c c c c }
    \toprule
    % \multirow{2}{*}{\textbf{Method}}
    % \multirow{2}{*}{\textbf{Method}} & \multirow{1}{*}{\textbf{Bit}}  & 
    % \multicolumn{2}{c}{\textbf{Set5}} & \multicolumn{2}{c}{\textbf{Set14}} & \multicolumn{2}{c}{\textbf{B100}} & \multicolumn{2}{c}{\textbf{Urban100}} & \multicolumn{2}{c}{\textbf{Manga109}}\\
    \multirow{2}{*}{\textbf{Method}} & \multirow{2}{*}{\textbf{Scale}} & Bit& \multicolumn{3}{c}{\textbf{Set5}}& \multicolumn{3}{c}{\textbf{Set14}}&\multicolumn{3}{c}{\textbf{B100}}&\multicolumn{3}{c}{\textbf{Urban100}}&\multicolumn{3}{c}{\textbf{Manga109}}\\
    \cmidrule{4-18}
    &&(w/a)&PSNR$\uparrow$&SSIM$\uparrow$&LPIPS$\downarrow$&PSNR$\uparrow$&SSIM$\uparrow$&LPIPS$\downarrow$&PSNR$\uparrow$&SSIM$\uparrow$&LPIPS$\downarrow$&PSNR$\uparrow$&SSIM$\uparrow$&LPIPS$\downarrow$&PSNR$\uparrow$&SSIM$\uparrow$&LPIPS$\downarrow$\\
    \midrule
    \rowcolor{gray!20}
    MambaIR \cite{guo2024mambair}&$2\times$ &32/32 &38.60 &0.9628 &0.0515 &34.69 &0.9261 &0.0842 &32.58 &0.9048 &0.1349 &34.15 &0.9446 &0.0470 &40.28 &0.9806 &0.0181\\
    % \hdashline
    Quamba* \cite{chiang2025quamba}&$2\times$ &8/8 &38.47&0.9622& 0.0521&34.50&0.9253&0.0840&32.53&0.9042&0.1350&33.97&0.9437&0.0472&39.99&0.9803 &0.0185        \\
    Quamba2* \cite{chiang2025quamba2}&$2\times$ &8/8&38.47&0.9623& 0.0518 &34.51& 0.9253& 0.0838  &32.53&0.9043& 0.1341 &33.96&0.9437& 0.0471 &39.99&0.9802 &0.0184 \\
    MambaQuant* \cite{xu2025mambaquant}&$2\times$ &8/8&38.52&0.9623&0.0509&34.65&0.9257&0.0827&32.56&0.9044&0.1334&34.11&0.9442&0.0464&40.16&0.9801&0.0180\\
    % \hdashline\\
    \midrule
    Quamba*\cite{chiang2025quamba}&$2\times$ &4/4 &29.17&0.7399&0.2252&27.62&0.7157& 0.1939 &27.26&0.6911& 0.2588&25.35&0.6818&0.2542 &27.89&0.7225&0.2194 \\
    Quamba2*\cite{chiang2025quamba2}&$2\times$ &4/4&31.98&0.8394& 0.1058&19.99&0.4329&0.5658&24.73&0.5484& 0.4495&27.18&0.7902&0.1749 &29.61& 0.8151&0.1489 \\
    % MambaQuant* &$2\times$ &4/4&31.48&0.8463&29.89&0.8438&30.73&0.8608&26.53&0.7938&30.94&0.8721\\
    MambaQuant* \cite{xu2025mambaquant}&$2\times$ &4/4 &19.37&0.3417&0.6493&17.20&0.2466&0.7137& 21.21&0.3625&0.6159& 17.18&0.2664&0.7549&14.49& 0.2044&0.7103  \\
    LSQ  \cite{esser2019learned}& $2\times$ & 4/4 & \textcolor{blue}{38.28} & \textcolor{blue}{0.9614} &\textcolor{red}{0.0532}& \textcolor{blue}{34.33} & \textcolor{blue}{0.9240} &\textcolor{red}{0.0846}  & \textcolor{blue}{32.43} & \textcolor{blue}{0.9030}  &\textcolor{red}{0.1356} & \textcolor{blue}{33.42} & \textcolor{blue}{0.9400} &\textcolor{blue}{0.0499}  & \textcolor{blue}{39.58}  & \textcolor{blue}{0.9685} &\textcolor{red}{0.0175 } \\
    QuantSR \cite{qin2023quantsr}& $2\times$ & 4/4 & 38.20 & 0.9609 &0.0542 & 34.13 & 0.9223 &0.0872  & 32.33 & 0.9014 &0.1372 & 32.73 & 0.9340 &0.0511  & 39.31 & 0.9781 & 0.0183\\
    Q-MambaIR (ours) & $2\times$ & 4/4 & \textcolor{red}{38.37} & \textcolor{red}{0.9617} &\textcolor{blue}{0.0533}& \textcolor{red}{34.44} & \textcolor{red}{0.9242} &\textcolor{blue}{0.0858}& \textcolor{red}{32.45} & \textcolor{red}{0.9031} &\textcolor{blue}{0.1365}& \textcolor{red}{33.64} & \textcolor{red}{0.9414} &\textcolor{red}{0.0497}& \textcolor{red}{39.76} & \textcolor{red}{0.9796} &\textcolor{blue}{0.0182} \\
    % Q-MambaIR2 \\
    \midrule
    % Quamba &$2\times$ &2/2 &----\\
    % MambaQuant &$2\times$ &2/2\\
    LSQ \cite{esser2019learned}& $2\times$ & 2/2 & 36.82 & 0.9555 &0.0614& 32.52 & 0.9076 &0.1084 & 31.35 & 0.8884 &0.1658  & 29.40 & 0.8951 & 0.1067 & 35.98 & 0.9789 &0.0307 \\
    QuantSR \cite{qin2023quantsr}& $2\times$ & 2/2 & \textcolor{blue}{37.67} & \textcolor{blue}{0.9592}  &\textcolor{blue}{0.0604}& \textcolor{blue}{33.25} & \textcolor{blue}{0.9152} &\textcolor{blue}{0.0972}& \textcolor{blue}{31.94} & \textcolor{blue}{0.8965} &\textcolor{blue}{0.1522} & \textcolor{blue}{31.35} & \textcolor{blue}{0.9212} &\textcolor{blue}{0.0812} & \textcolor{blue}{38.10} & \textcolor{blue}{0.9757}& \textcolor{blue}{0.0255}\\
    Q-MambaIR (ours)  & $2\times$ & 2/2 & \textcolor{red}{37.77} & \textcolor{red}{0.9598} &\textcolor{red}{0.0563}& \textcolor{red}{33.50} & \textcolor{red}{0.9169} &\textcolor{red}{0.0957}& \textcolor{red}{32.06} & \textcolor{red}{0.8983} &\textcolor{red}{0.1470}& \textcolor{red}{31.88} & \textcolor{red}{0.9269} &\textcolor{red}{0.0646}& \textcolor{red}{38.25} & \textcolor{red}{0.9764} &\textcolor{red}{0.0235}\\
    % Q-MambaIR2 \\
    % \midrule
    % MambaIR &$3\times$ &32/32 &35.08 &0.9323 &30.99 &0.8536 &29.51 &0.8157 &29.93 &0.8841 &35.43 &0.9546\\
    % Quamba &$3\times$ &4/4 \\
    % MambaQuant &$3\times$ &4/4\\
    % QuantSR &$3\times$ &4/4\\
    % Ours &$3\times$ &4/4\\ 
    % \midrule
    % Quamba &$3\times$ &2/2\\
    % MambaQuant &$3\times$ &2/2\\
    % QuantSR &$3\times$ &2/2\\
    % Ours &$3\times$ &2/2 \\
    \midrule
    \rowcolor{gray!20}
    MambaIRv2 \cite{guo2025mambairv2} &$2\times$ &32/32 & 38.65 &0.9631 & 0.0512 &34.88 &0.9275 &0.0811 &32.62 &0.9052  &0.1322  &34.49 &0.9468 &0.0445&40.42 &0.9810&0.0174  \\
    % \hdashline
    % Mambaquant &$2\times$ &4/4\\
    LSQv2 \cite{esser2019learned} &$2\times$ &4/4 &38.12 &0.9609 &0.0546 &33.82 &0.9201 &0.0889 &32.31 &0.9013 &0.1365&32.86&0.9350&0.0546&39.12&0.9782&0.0225 \\
    QuantSRv2 \cite{qin2023quantsr} &$2\times$ &4/4 &\textcolor{blue}{38.14} &\textcolor{blue}{0.9610} &\textcolor{red}{0.0540} &\textcolor{blue}{33.88} &\textcolor{blue}{0.9211} &\textcolor{blue}{0.0876} &\textcolor{blue}{32.35} &\textcolor{blue}{0.9019 }&\textcolor{blue}{0.1368} &\textcolor{blue}{33.04}&\textcolor{blue}{0.9370}&\textcolor{blue}{0.0536} & \textcolor{blue}{39.19}&\textcolor{blue}{0.9784}	&\textcolor{blue}{0.0223} \\
    Q-MambaIRv2 (ours)  &$2\times$ &4/4 &\textcolor{red}{ 38.33} &\textcolor{red}{0.9616} &\textcolor{blue}{0.0543}   &\textcolor{red}{34.18} &\textcolor{red}{0.9229} &\textcolor{red}{0.0871}&\textcolor{red}{32.45} &\textcolor{red}{0.9032} &\textcolor{red}{0.1343}&\textcolor{red}{33.52} &\textcolor{red}{ 0.9408} &\textcolor{red}{0.0506} &\textcolor{red}{38.91} &\textcolor{red}{0.9776} &\textcolor{red}{0.0215}\\
    \midrule
    LSQv2 \cite{esser2019learned} &$2\times$ &2/2 &\textcolor{blue}{36.71} &\textcolor{blue}{0.9547} &0.0654&\textcolor{blue}{32.45} &\textcolor{blue}{0.9068} &0.1130 &\textcolor{blue}{31.33} &\textcolor{blue}{0.8875} & 0.1721 &\textcolor{blue}{29.40} &\textcolor{blue}{0.8943} & 0.1189&\textcolor{blue}{35.74} &\textcolor{blue}{0.9673} &0.0364 \\
    QuantSRv2 \cite{qin2023quantsr} &$2\times$ &2/2 &36.42 &0.9525 &\textcolor{red}{0.0488}&32.32&0.9050 &\textcolor{red}{0.0842}&31.22 & 0.8856& \textcolor{red}{0.1303} &29.78 &0.9005 &\textcolor{red}{0.0761}&35.37 &0.9652&\textcolor{red}{0.0251} \\
    Q-MambaIRv2 (ours)   &$2\times$ &2/2  &\textcolor{red}{37.26} &\textcolor{red}{0.9573}&\textcolor{blue}{0.0610}  &\textcolor{red}{32.85} &\textcolor{red}{0.9108} &\textcolor{blue}{0.1066}&\textcolor{red}{31.64} &\textcolor{red}{ 0.8923} &\textcolor{blue}{0.1596} &\textcolor{red}{30.39} &\textcolor{red}{ 0.9095 }&\textcolor{blue}{0.0926} &\textcolor{red}{37.10} &\textcolor{red}{0.9727}&\textcolor{blue}{0.0279}\\
    % \rowcolor{gray!20}
    % MambaIR \cite{guo2024mambair} &$3\times$ &32/32 \\
    % % \hdashline
    % Quamba* \cite{chiang2025quamba} &$3\times$ &8/8  \\
    % MambaQuant* \cite{xu2025mambaquant} &$3\times$ &8/8 \\
    % % \hdashline
    % Quamba* \cite{chiang2025quamba} &$3\times$ &4/4  \\
    % MambaQuant* \cite{xu2025mambaquant}&$3\times$ &4/4  \\
    % LSQ \cite{esser2019learned} & $3\times$ & 4/4  \\
    % QuantSR \cite{qin2023quantsr} & $3\times$ & 4/4  \\
    % Q-MambaIR & $3\times$ & 4/4 &   \\
    % \midrule
    % % Quamba &$4\times$ &2/2 &----\\
    % % MambaQuant &$4\times$ &2/2\\
    % LSQ \cite{esser2019learned} & $3\times$&2/2\\
    % QuantSR \cite{qin2023quantsr} & $3\times$ & 2/2  \\
    % Q-MambaIR & $3\times$ & 2/2  \\
    \midrule
    \midrule
    \rowcolor{gray!20}
    MambaIR \cite{guo2024mambair}&$4\times$ &32/32 &33.03 &0.9046 &0.1646 &29.20 &0.7961 & 0.2680 &27.98 &0.7503 &0.3556&27.68 &0.8287& 0.1833&32.32 &0.9272&0.0912\\
    % \hdashline
    Quamba* \cite{chiang2025quamba}&$4\times$ &8/8&32.99&0.9038&0.1651 &29.14&0.7948&0.2691&27.95&0.7492& 0.3576&27.58&0.8268& 0.1851&32.10&0.9255&0.0920 \\
    Quamba2* \cite{chiang2025quamba2}&$4\times$ &8/8 &33.00&0.9039& 0.1655&29.15& 0.7947&0.2687&27.95&0.7493&0.3570&27.59& 0.8268	&0.1851&32.11&0.9256&0.0918\\
    MambaQuant*\cite{xu2025mambaquant} &$4\times$ &8/8&32.98& 0.9036&0.1608 &29.18&0.7953&0.2648 &27.97&0.7497&0.3527 &27.66&0.8279&0.1823 &32.20&0.9260&0.0893  \\
    % \hdashline
    \midrule
    Quamba* \cite{chiang2025quamba}&$4\times$ &4/4 &26.19&0.6912&0.5372 &25.53&0.6342&0.5709 &26.15&0.6448&0.6032 &22.04&0.5409&0.6024 &24.70&0.6811&0.4877 \\
    Quamba2* \cite{chiang2025quamba2} &$4\times$ &4/4 &31.98&0.8394&0.1058&19.99&0.4329&0.5658&24.73&0.5484&0.4495&27.18& 0.7902 &0.1749  &29.61&0.8151&0.1488\\
    % MambaQuant* &$4\times$ &4/4 &26.18&0.6904&25.54&0.6350&26.15&0.6448&22.04&0.5409&24.70& 0.6811\\
    MambaQuant*\cite{xu2025mambaquant} &$4\times$ &4/4 &19.36&0.3397&0.6445&17.71&0.2559	&0.7130&21.30&0.3664&0.6144&17.17&0.2657&0.7555&14.49&0.2036&0.7104\\
    LSQ \cite{esser2019learned}& $4\times$ & 4/4 & 32.54 & 0.8988 &0.1724& 28.85 & 0.7877 &0.2791& 27.75 & 0.7426 &0.3673& 26.80 & 0.8089 &0.2092& 31.17 & 0.9153& 0.1011  \\
    QuantSR \cite{qin2023quantsr}& $4\times$ & 4/4 & \textcolor{blue}{32.68} & \textcolor{blue}{0.8998} &\textcolor{blue}{0.1683}& \textcolor{blue}{28.79} & \textcolor{blue}{0.7868} &\textcolor{blue}{0.2742}& \textcolor{blue}{27.75} & \textcolor{blue}{0.7430} &\textcolor{blue}{0.3640}& \textcolor{blue}{26.19} & \textcolor{blue}{0.7912} &\textcolor{blue}{0.2081}& \textcolor{blue}{30.74} & \textcolor{blue}{0.9120} &\textcolor{blue}{0.0997} \\
    Q-MambaIR (ours)  & $4\times$ & 4/4 & \textcolor{red}{32.79} & \textcolor{red}{0.9016} &\textcolor{red}{0.1682}& \textcolor{red}{29.04} & \textcolor{red}{0.7924} &\textcolor{red}{0.2714}& \textcolor{red}{27.88} & \textcolor{red}{0.7465} &\textcolor{red}{0.1917}& \textcolor{red}{27.26} & \textcolor{red}{0.8193} &\textcolor{red}{0.0948}& \textcolor{red}{31.69} & \textcolor{red}{0.9208} &\textcolor{red}{0.0948}\\
    % Q-MambaIR2 \\
    \midrule
    % Quamba &$4\times$ &2/2 &----\\
    % MambaQuant &$4\times$ &2/2\\
    LSQ \cite{esser2019learned} & $4\times$ & 2/2 & 30.77 & 0.8712 & 0.2243 & 27.70 & 0.7595 &0.3379 & 27.00 & 0.7164 &0.4352 & 24.64 & 0.7327 &0.3563 & 27.81 & 0.8648 &0.2003 \\
    QuantSR \cite{qin2023quantsr}& $4\times$ & 2/2 & \textcolor{blue}{31.76} & \textcolor{blue}{0.8888} & \textcolor{blue}{0.1769}&  \textcolor{blue}{28.29} & \textcolor{blue}{0.7743} & \textcolor{blue}{0.2919}& \textcolor{blue}{27.38} & \textcolor{blue}{0.7296} & \textcolor{blue}{0.3829}&  \textcolor{blue}{25.60} & \textcolor{blue}{0.7704} & \textcolor{blue}{0.2486}& \textcolor{blue}{29.64} & \textcolor{blue}{0.8970} & \textcolor{blue}{0.1167} \\
    Q-MambaIR (ours)  & $4\times$ & 2/2 & \textcolor{red}{32.02} & \textcolor{red}{0.8920} &\textcolor{red}{0.1744}& \textcolor{red}{28.48} & \textcolor{red}{0.7792} &\textcolor{red}{0.2847}& \textcolor{red}{27.54} & \textcolor{red}{0.7350} &\textcolor{red}{0.3732}& \textcolor{red}{26.00} & \textcolor{red}{0.7849} &\textcolor{red}{0.2288}& \textcolor{red}{30.17} & \textcolor{red}{0.9045} &\textcolor{red}{0.1092}\\
    \midrule
    \rowcolor{gray!20}
    MambaIRv2 \cite{guo2025mambairv2}&$4\times$ &32/32 & 33.15 &0.9059 &0.1633 &29.23 &0.7975 &0.2657  &  28.00&0.7511 &0.3545 &27.89&0.8344& 0.1796&32.57&0.9295&0.0901\\
    LSQv2 \cite{esser2019learned} &$4\times$ &4/4 &\textcolor{blue}{32.56} &0.8984&\textcolor{blue}{0.1693}&\textcolor{blue}{28.84}  &\textcolor{blue}{0.7872} &\textcolor{red}{0.2721}&\textcolor{blue}{27.74} &\textcolor{blue}{0.7422} &\textcolor{red}{0.3617} &\textcolor{blue}{ 26.71} &\textcolor{blue}{0.8070 } &\textcolor{blue}{0.2014} &\textcolor{blue}{ 31.05 }&\textcolor{blue}{0.9143} &\textcolor{red}{0.0980  }  \\
    QuantSRv2 \cite{qin2023quantsr} &$4\times$ &4/4 &32.54 & \textcolor{blue}{0.8987} &\textcolor{red}{0.1683} &28.82 &0.7867 &\textcolor{blue}{0.2742}  &27.73 &0.7419 &0.3640  &26.63 &0.8046 &0.2081 &31.01 &0.9141 &0.0997  \\
    Q-MambaIRv2 (ours)   &$4\times$ &4/4 &\textcolor{red}{32.63} &\textcolor{red}{0.8994} &0.1703 &\textcolor{red}{28.92} &\textcolor{red}{0.7887} &0.2761&\textcolor{red}{27.78} &\textcolor{red}{0.7432} &\textcolor{blue}{0.3620} &\textcolor{red}{26.93}&\textcolor{red}{0.8123}&\textcolor{red}{0.2006}&\textcolor{red}{31.35}&\textcolor{red}{0.9176} &\textcolor{blue}{0.0981} \\
    \midrule
    % Quamba &$4\times$ &2/2 &l-ing\\
    % Mambaquant &$4\times$ &2/2\\
    LSQv2 \cite{esser2019learned} & $4\times$ & 2/2 & \textcolor{blue}{23.88} &\textcolor{blue}{  0.5103}  &\textcolor{blue}{0.4223} &\textcolor{blue}{22.42} &\textcolor{blue}{0.4271} &\textcolor{blue}{0.4398} &\textcolor{blue}{22.96} &\textcolor{blue}{0.4333} & \textcolor{blue}{0.4742} &\textcolor{blue}{20.77}&\textcolor{blue}{0.4104}&\textcolor{blue}{0.5083}&\textcolor{blue}{21.91}&\textcolor{blue}{0.4693}& \textcolor{blue}{0.4403} \\
    QuantSRv2 \cite{qin2023quantsr} & $4\times$ & 2/2 &22.27 & 0.4019 &0.4839& 20.17 & 0.2839&0.5385   & 20.80 & 0.3007 &0.5689& 18.38&0.2581 &0.5815& 19.78 & 0.3074&0.5056 \\
    Q-MambaIRv2 (ours)    & $4\times$ & 2/2 & \textcolor{red}{30.42} & \textcolor{red}{ 0.8630} & \textcolor{red}{0.2287}& \textcolor{red}{27.43} & \textcolor{red}{0.7523} &\textcolor{red}{0.3421} & \textcolor{red}{26.85} & \textcolor{red}{0.7106} &\textcolor{red}{0.4386} & \textcolor{red}{24.33} & \textcolor{red}{0.7176} &\textcolor{red}{0.3605} & \textcolor{red}{27.09} & \textcolor{red}{0.8468} & \textcolor{red}{0.2071}\\
    % Q-MambaIR2 \\
    \bottomrule
    \end{tabular}
  \end{adjustbox} 
  \vspace{-3mm}
\end{table*}
\begin{table}
  \centering
  \caption{Quantitative comparison on \textbf{JPEG compression artifact reduction} with a quality factor of 30.}
  \begin{adjustbox}{width=\linewidth}
  \begin{tabular}{@{} l c c c c c}
    \toprule
    % \multirow{2}{*}{\textbf{Method}}
    % \multirow{2}{*}{\textbf{Method}} & \multirow{1}{*}{\textbf{Bit}}  & 
    % \multicolumn{2}{c}{\textbf{Set5}} & \multicolumn{2}{c}{\textbf{Set14}} & \multicolumn{2}{c}{\textbf{B100}} & \multicolumn{2}{c}{\textbf{Urban100}} & \multicolumn{2}{c}{\textbf{Manga109}}\\
    Method &Bit& \multicolumn{2}{c}{\textbf{Classic5}}& \multicolumn{2}{c}{\textbf{LIVE1}}\\
    \cmidrule{3-6}
    &(w/a)&PSNR&SSIM&PSNR&SSIM\\
    % & &(w/a)&$\sigma$=15&$\sigma$=25&$\sigma$=50&$\sigma$=15&$\sigma$=25&$\sigma$=50&$\sigma$=15&$\sigma$=25&$\sigma$=50&$\sigma$=15&$\sigma$=25&$\sigma$=50\\
    \midrule
    \rowcolor{gray!20}
    MambaIR \cite{guo2024mambair}  &32/32 &33.74&0.8965&33.72&0.9178\\
    \midrule
    Quamba \cite{chiang2025quamba}  &8/8 &33.74&0.8961 &33.73&0.9178\\
    MambaQuant \cite{xu2025mambaquant}&8/8&32.24&0.8722&32.24&0.8928\\
    % \hdashline
    \midrule
    Quamba \cite{chiang2025quamba} &4/4 &31.12&0.8246&31.06&0.8376\\
    MambaQuant \cite{xu2025mambaquant}&4/4&24.46&0.5045&24.61&0.5161\\
    LSQ \cite{esser2019learned} &4/4 &\textcolor{blue}{33.51} &\textcolor{blue}{0.8937} &\textcolor{blue}{33.54} &\textcolor{blue}{0.9156} \\
    QuantSR \cite{qin2023quantsr}  &4/4 &33.29 &0.8907 &33.34 &0.9131 \\
    Q-MambaIR (ours)  &4/4 &\textcolor{red}{33.63} &\textcolor{red}{0.8950} &\textcolor{red}{33.63} &\textcolor{red}{0.9166} \\
    \midrule
    % Quamba &30 &2/2\\
    % Mambaquant &30 &2/2\\
    % \hdashline
    LSQ \cite{esser2019learned} &2/2 &32.26 &0.8767 &32.41 &0.8988 \\
    QuantSR \cite{qin2023quantsr} &2/2 &\textcolor{blue}{32.80} &\textcolor{blue}{0.8843} &\textcolor{blue}{32.88} &\textcolor{blue}{0.9073} \\
    Q-MambaIR (ours)  &2/2 &\textcolor{red}{32.86} &\textcolor{red}{0.8852} &\textcolor{red}{32.92} &\textcolor{red}{0.9080} \\
    \bottomrule
    \end{tabular}
  \end{adjustbox}
  \label{tab:jpeg}
  \vspace{-4mm}
\end{table}
\begin{table}
  % \vspace{2mm}
  \centering
  \caption{Quantitative comparison on \textbf{real image denoising} task.}
  \begin{adjustbox}{width=\linewidth}
  \begin{tabular}{@{} l c c c c c}
    \toprule
    % \multirow{2}{*}{\textbf{Method}}
    % \multirow{2}{*}{\textbf{Method}} & \multirow{1}{*}{\textbf{Bit}}  & 
    % \multicolumn{2}{c}{\textbf{Set5}} & \multicolumn{2}{c}{\textbf{Set14}} & \multicolumn{2}{c}{\textbf{B100}} & \multicolumn{2}{c}{\textbf{Urban100}} & \multicolumn{2}{c}{\textbf{Manga109}}\\
    Method &Bit& \multicolumn{2}{c}{\textbf{SIDD}}& \multicolumn{2}{c}{\textbf{DND}}\\
    % \cmidrule{3-6}
    &(w/a)&PSNR&SSIM&PSNR&SSIM\\
    % & &(w/a)&$\sigma$=15&$\sigma$=25&$\sigma$=50&$\sigma$=15&$\sigma$=25&$\sigma$=50&$\sigma$=15&$\sigma$=25&$\sigma$=50&$\sigma$=15&$\sigma$=25&$\sigma$=50\\
    \midrule
    \rowcolor{gray!20}
    MambaIR \cite{guo2024mambair}  &32/32 &39.89 &0.960&40.04&0.956\\
    \midrule
    % Quamba \cite{chiang2025quamba}  &8/8 \\
    % Mambaquant \cite{xu2025mambaquant}&8/8\\
    % \hdashline
    % Quamba \cite{chiang2025quamba} &4/4 \\
    % Mambaquant \cite{xu2025mambaquant}&4/4\\
    LSQ \cite{esser2019learned} &4/4 &39.24 &0.9555 &\textcolor{blue}{39.33}&	\textcolor{blue}{0.9507}\\
    QuantSR \cite{qin2023quantsr}  &4/4 &\textcolor{blue}{39.26}&\textcolor{blue}{0.9507}&39.27&0.9488\\
    Q-MambaIR (ours)  &4/4 &\textcolor{red}{39.50}&\textcolor{red}{0.9559}&\textcolor{red}{39.46}&\textcolor{red}{0.9571}\\
    \midrule
    % Quamba &30 &2/2\\
    % Mambaquant &30 &2/2\\
    % \hdashline
    LSQ \cite{esser2019learned} &2/2  &38.39 &0.9499&\textcolor{blue}{39.06}&\textcolor{blue}{0.9474} \\
    QuantSR \cite{qin2023quantsr} &2/2 &\textcolor{blue}{38.75}&\textcolor{blue}{0.9508}&38.88&0.9466\\
    Q-MambaIR (ours) &2/2 &\textcolor{red}{38.87}&\textcolor{red}{0.9530}&\textcolor{red}{39.33}&\textcolor{red}{0.9501}\\
    \bottomrule
    \end{tabular}
  \end{adjustbox}
  \label{tab:rDN}
  \vspace{-4mm}
\end{table}

 \subsubsection{Proposed Quantization Baselines}
 We quantized two existing SSM-based IR models, i.e., MambaIR \cite{guo2024mambair} and MambaIRv2 \cite{guo2025mambairv2}, with our proposed method. We quantize the body part of the network in all comparison models at low bitwidths (e.g., 8-, 4-, or 2-bit). For fair comparison, the implementations of comparison methods follow the official codes and are trained with the same settings as ours. For PTQ methods, we follow the mamba quantization strategies MambaQuant \cite{xu2025mambaquant}, Quamba \cite{chiang2025quamba} and Quamba2 \cite{chiang2025quamba2}. For QAT methods, we quantize SSM with QuantSR \cite{qin2023quantsr} and LSQ \cite{esser2019learned}. We denote $w$-bit weight and $a$-bit activation as $w/a$.

\subsubsection{Training Strategy}
According to previous works \cite{chen2023activating,guo2024mambair}, we perform data augmentation by applying horizontal flips and random rotations of $90^\circ$, $180^\circ$, and $270^\circ$. Additionally, we cropped the original images into 64×64 patches for image SR and 128×128 patches for image denoising during training. The models are trained for 20k iterations, with each training batch consisting of 8 image patches for image SR and 4 for image denoising. The Adam optimizer is utilized. The learning rate is initially set to
$2 \times 10^{-4}$ for lightweight SR and $1 \times 10^{-4}$ for other tasks, and is then halved when the training iteration reaches specific milestones.
Our Q-MambaIR model is trained with 2 NVIDIA A100 GPUs. More training details are available in \textit{Supplementary Material}.

\subsection{Comparison on State-of-the-Art Methods}\label{sec:comp1}
We select MambaIR \cite{guo2024mambair} (\ie, 82,045K (2x) and 82,637K (4x) Params) and MambaIRv2 (\ie, 92,812K (2x) and 93,450K (4x) Params) \cite{guo2025mambairv2} as our backbone and compare our method with recent quantization methods. For PTQ, we evaluate MambaQuant \cite{xu2025mambaquant}, Quamba \cite{chiang2025quamba}, Quamba2 \cite{chiang2025quamba2}, and PTQ4VM \cite{cho2025ptq4vm}. For QAT, we evaluate QuantSR \cite{qin2023quantsr} and LSQ \cite{esser2019learned}. We further validate the generalizability of Q-MambaIR across additional SSM-based (EAMamba~\cite{lin2025eamamba}, VmambaIR~\cite{shi2025vmambair}, MaIR~\cite{li2025mair}), CNN-based (EDSR~\cite{lim2017enhanced}), and Transformer-based (SwinIR~\cite{liang2021swinir}) IR architectures in the \textit{Supplementary Material}.
% We select MambaIR \cite{guo2024mambair} (\ie, 82,045K (2x) and 82,637K (4x) Params) and MambaIRv2 (\ie, 92,812K (2x) and 93,450K (4x) Params) \cite{guo2025mambairv2} as our backbone and compare our method with recent quantization methods. We select two mamba quantization PTQ methods, i.e. MambaQuant \cite{xu2025mambaquant}, Quamba \cite{chiang2025quamba} and Quamba2 \cite{chiang2025quamba2}. For QAT methods, we evaluate the performance of QuantSR \cite{qin2023quantsr} and LSQ \cite{esser2019learned}. We also do further experiments on CNN-based and SSM-Transformer IR architectures in \textit{Supplementary Material}.

\subsubsection{Image Restoration} Our proposed model consistently outperforms the baseline method across all tasks in IR, including image super-resolution, real image denoising, Gaussian  color image denoising, and JPEG compression artifact reduction. \\
\noindent\textbf{Image Super Resolution.} Tab.~\ref{tab:classic} provides a comparison of different methods on the classic image SR. Our 4-bit Q-MambaIR achieves comparable or superior PSNR/SSIM scores to the 8-bit Quamba and MambaQuant with a scale of $2\times$. In the 4-bit case, our Q-MambaIR achieves PSNR/SSIM values that are 0.21 dB/0.0014 ($2\times$) and 0.46 dB/0.0104 ($4\times$) higher than LSQ on Urban100. With a 2-bit setting, our Q-MambaIR outperforms QuantSR by 0.53 dB/0.0057 ($2\times$) and 0.4 dB/0.0145 ($4\times$) on Urban100. \\
\noindent\textbf{JPEG Compression Artifact Reduction. }We further evaluate the robustness of Q-MambaIR on the task of JPEG compression artifact reduction with a quality factor of 30. The results are summarized in \cref{tab:jpeg}. As shown in the table, on the Classic5 dataset, our method outperforms the existing quantization baseline by \textbf{0.12 dB} in the 4-bit setting. This demonstrates that Q-MambaIR not only maintains high restoration quality on clean inputs but also generalizes effectively to compressed and degraded images.\\
\noindent\textbf{Real Image Denoising.} We further evaluate Q-MambaIR on the real image denoising task using the SIDD and DND datasets, with results reported in Tab.~\ref{tab:rDN}. Under 4-bit quantization, our method achieves 39.50\,dB on SIDD, surpassing QuantSR~\cite{qin2023quantsr} by 0.24\,dB and approaching the full-precision MambaIR (39.89\,dB) within 0.4\,dB. At the more challenging 2-bit setting, Q-MambaIR maintains a clear advantage over both LSQ and QuantSR on SIDD and DND, demonstrating strong robustness against real-world noise patterns under aggressive quantization. Extended results on Gaussian color image denoising are provided in the \textit{Supplementary Material}.
% \noindent\textbf{Gaussian Denoising.} A representative case of Gaussian  color image denoising at a noise level of 25 is provided in Tab.~\ref{tab:DN}. In particular, under 4-bit quantization on the Urban100 dataset, our approach surpasses the performance of QuantSR \cite{qin2023quantsr} by up to 0.48 dB in PSNR. At lower bitwidth, our Q-mambaIR has a significant improvement compared to the previous technique and narrows the gap between low-bit models and full-precision models. Our method provides a possible SR solution for ultra-low bit quantization. Other results for extended experiments are provided in \textit{Supplementary Material}.\\

\begin{table}
    \caption{Model complexity of Q-MambaIR ($2\times$ SR)}
    \begin{adjustbox}{width=.5\textwidth}
    % \vspace{-2mm}
    \label{tab:param}
  \centering
  \small
  \begin{tabular}{@{}l c c c c}
    \toprule
    Method & Bits &\#Params 
    &Flops&Urban100  \\
    % \cmidrule{5-6}
    &(w/a)&(↓ Ratio)&(↓ Ratio)\\
    \midrule
    \rowcolor{gray!20}
    MambaIR-light & 32/32 &859.32{} K&88.02 G &32.92  \\
    \midrule
    Q-MambaIR (ours) & 4/4  &114.70 
 (↓86.7\%)  &15.71 (↓82.2\%)&32.43\\
    Q-MambaIR (ours) & 2/2  &80.67 (↓90.6\%)  &10.54 (↓88.0\%)&30.81\\
    \midrule
    \rowcolor{gray!20}
    MambaIR & 32/32 &13.072 M &1.334 T&34.15  \\
    \midrule
    Q-MambaIR (ours) & 4/4  & 3.67 (↓71.9\%)  & 0.38 (↓71.5\%) &33.64\\
    Q-MambaIR (ours) & 2/2  &2.99 (↓77.1\%)& 0.31 (↓77.8\%)&31.88\\
    \bottomrule
    \end{tabular}
  \end{adjustbox}
  \vspace{-3mm}
\end{table}

\subsubsection{Model Complexity} With ultra-low bitwidth quantization, the quantized model can achieve very impressive compression ratios. In Tab.~\ref{tab:param}, we
present the compression ratio and speedup in terms
of model size (Params) and number of operations (Ops), respectively. The reported reduction in parameters is based on an equivalent conversion from 32-bit parameter size. The calculation follows the methods used in \cite{qin2023quantsr}. With a 2-bit setting, our Q-MambaIR achieves a compression ratio of approximately 90.6\% and an acceleration ratio of 88.0\% for parameters and operations in lightweight MambaIR.

\begin{table*}
   \caption{Ablation study of components in Q-MambaIR on lightweight image SR (2$\times$). PSNR (dB) is reported on four benchmarks.}
   % \vspace{-2mm}
   \label{tab:ablation}
    \centering
    \small
    % \begin{adjustbox}{width=\linewidth}
  % \begin{threeparttable}
  % \begin{adjustbox}{width=.5\linewidth}
    % \setlength{\tabcolsep}{1pt}
    \begin{tabular}{@{}l c c c c c c c c}
  % \begin{tabular}{@{}l c >{\centering\arraybackslash}p{1.5cm} >{\centering\arraybackslash}p{1.5cm} >{\centering\arraybackslash}p{1.5cm} >{\centering\arraybackslash}p{1.5cm} >{\centering\arraybackslash}p{1.5cm}}
    \toprule
    Method &Bits (w/a) & Latency / ms &\textbf{Set14} & \textbf{B100} & \textbf{Urban100} &\textbf{Manga109}\\
    \midrule
    \rowcolor{gray!20}
    MambaIR &32/32  & 198.21 &34.00 & 32.34 &32.92 &39.31 \\
    \midrule
    Baseline  & 4/4  &119.89 & 33.44$\pm$0.03 & 32.04$\pm$0.02& 31.37$\pm$0.03 & 38.15$\pm$0.04 \\
    % DLS  & 4/4  & 38.00 &33.76 &32.22 &32.42 &38.88\\
    DLS  & 4/4   &120.87 &33.76$\pm$0.02 &32.21$\pm$0.03 &32.42$\pm$0.03&38.88$\pm$0.03\\
    RFA  & 4/4  &110.64 &33.50$\pm$0.03 &32.06$\pm$0.03  &31.38$\pm$0.08 & 38.16$\pm$0.05 \\
     % RFA  & 4/4 &37.84 &33.40 &32.03  &31.31 & 38.12 \\
    \midrule
    Q-MambaIR  & 4/4  &119.74 & 33.78$\pm$0.02 & 32.22$\pm$0.04 &32.43$\pm$0.05&38.91$\pm$0.02\\
    \bottomrule
    \end{tabular}
    % \end{adjustbox}
    \begin{tablenotes} %[flushleft]
      \footnotesize
      \item Note: Each reported value is the average of the first five experimental runs.
    \end{tablenotes}
    \vspace{-3mm}
   %  \end{threeparttable}
   % \end{adjustbox}
\end{table*}

\begin{table}
% \vspace{2mm}
  \centering
  \caption{Ablation on different design choices of DLS on lightweight image SR (2$\times$). PSNR (dB) is reported.}
  \begin{adjustbox}{width =\linewidth}
  \begin{tabular}{@{}l c c >{\centering\arraybackslash}p{0.9cm} >{\centering\arraybackslash}p{0.9cm} >{\centering\arraybackslash}p{0.9cm} >{\centering\arraybackslash}p{1.3cm} >{\centering\arraybackslash}p{1.3cm}}
    \toprule
    % Method & Bit& \multicolumn{2}{c}{\textbf{Set5}}& \multicolumn{2}{c}{\textbf{Set14}}&\multicolumn{2}{c}{\textbf{B100}}&\multicolumn{2}{c}{\textbf{Urban100}}&\multicolumn{2}{c}{\textbf{Manga109}}\\
    Method & $\alpha$&$\beta$& \textbf{Set5}& \textbf{Set14}&\textbf{B100}&\textbf{Urban100}&\textbf{Manga109}\\
    % \cmidrule{3-7}
    % &(w/a)&PSNR&SSIM&PSNR&SSIM&PSNR&SSIM&PSNR&SSIM&PSNR&SSIM\\
    \midrule
    % MambaIR-light \cite{guo2024mambair} & &&38.16&34&32.34&32.92&39.31\\
    % \midrule
    % Baseline & 4/4 &37.88&33.44&32.04&31.37&38.15\\
    \rowcolor{gray!20}
    GROUP1&$\alpha_1$&$\beta_1$&37.98&33.72&32.21&32.36&38.86\\  GROUP2&$\alpha_1$&$\beta_2$&37.03&32.65&31.46&29.66&36.50\\  GROUP3&$\alpha_2$&$\beta_1$&37.71&33.17&31.91&31.01&37.93\\ GROUP4&$\alpha_2$&$\beta_2$&37.68&33.17&31.90&30.98&37.92\\
    \bottomrule
  \end{tabular}
  \end{adjustbox}
  \label{tab:INIT}
  \vspace{-3mm}
\end{table}
\subsection{Ablation Study}
\label{sec:ablation}
% \yawli{Check Sec.~\ref{subsec:rfa} for more ablation study.} 

To demonstrate the effectiveness of the techniques employed in our Q-MambaIR, we conduct ablation studies on DLS and RFA. We adopt MambaIR as the image restoration backbone and train it for 20K iterations with 4-bit and 2-bit quantization settings. To establish a quantization baseline, we utilize the vanilla quantization method, QuantSR \cite{qin2023quantsr}. 

\subsubsection{Components in Q-MambaIR.} 
We incorporate DLS and/or RFA into MambaIR and quantize it to 4-bit. The experimental results are summarized in Tab.~\ref{tab:ablation}, where we report the PSNR values on four test benchmarks. The vanilla backbone demonstrates the feasibility of low-bit quantization. As shown in Tab.~\ref{tab:ablation}, the incorporation of DLS alone improves the accuracy by 1.05\,dB on Urban100 compared to the baseline, confirming the importance of input-adaptive quantization for SSM activations. Using RFA alone yields marginal gains (+0.01\,dB), as it cannot fully recover the truncation loss caused by static activation ranges. When DLS and RFA are combined, they complement each other: DLS addresses activation quantization while RFA refines weight discretization, achieving superior detail preservation with a PSNR of 32.43\,dB.

\subsubsection{Fine-grained Quantization Design Choices.}
To further justify the design of DLS and RFA, we compare them against representative alternatives for activation and weight quantization, respectively. Tab.~\ref{tab:quantizer} isolates the activation quantizer (with RFA fixed for weights): DLS outperforms all learned-fixed methods by +1.07$\sim$+4.41\,dB, confirming that input-adaptive range prediction is essential for SSM's shifting activation distributions. Tab.~\ref{tab:quantizer} isolates the weight quantizer (with DLS fixed for activations): RFA outperforms STE and its variants by +0.63\,dB, validating the benefit of adaptive rounding with accurate gradient estimation for SSM weights.
\begin{table}[h]
\centering
\caption{Fine-grained ablation on the design choices of DLS (activation) and RFA (weight). PSNR (dB) is reported on Urban100 for lightweight SR (2$\times$). Models are trained for 10k iterations.}
\label{tab:quantizer}
\noindent
\begin{minipage}[t]{0.48\linewidth}
\centering
\scriptsize
(a) Activation Quantization\\[1mm]
\begin{tabular}{@{}lcc@{}}
\toprule
Method & Type & PSNR \\
\midrule
QuantSR & static $\alpha$, RLQ & 31.17 \\
LSQ+~\cite{bhalgat2020lsqplus} & static $\alpha,\beta$ & 30.78 \\
PAMS~\cite{li2020pams} & content-aware & 27.83 \\
\rowcolor{gray!15}
\textbf{DLS} & input-adaptive & \textbf{32.24} \\
\bottomrule
\end{tabular}
\end{minipage}
\hfill
\begin{minipage}[t]{0.48\linewidth}
\centering
\scriptsize
(b) Weight Quantization\\[1mm]
\begin{tabular}{@{}lcc@{}}
\toprule
Method & Backward & PSNR \\
\midrule
STE & identity & 30.37 \\
Soft STE & tanh-based & 30.40 \\
NU2U~\cite{liu2022nonuniform} & piecewise & 30.39 \\
\rowcolor{gray!15}
\textbf{RFA} & adaptive piecewise & \textbf{31.03} \\
\bottomrule
\end{tabular}
\end{minipage}
\vspace{-4mm}
\end{table}
\subsubsection{Computational Overhead.} 
We conduct latency evaluations on the \textit{Jetson Orin} platform to assess the practical deployment cost of the proposed components. As reported in Tab.~\ref{tab:ablation}, Q-MambaIR maintains inference latency comparable to the baseline 4-bit model (119.89\,ms vs.\ 119.74\,ms), demonstrating negligible overhead. This is because DLS relies on lightweight global reduction operations: it incurs approximately $14N$ FLOPs per activation tensor ($N$ elements), corresponding to $\sim$173\,MFLOPs per SS2D block, accounting for less than 0.1\% of the selective scan computation. RFA applies quantization offline and introduces no additional inference cost. Detailed analysis is in the \textit{Supplementary Material}.

\subsubsection{Initialization of Weights in DLS.}
We further investigate various designs for quantization range initialization in DLS. Specifically, we initialize $\alpha$ and $\beta$ to align the initial quantization range with intuitive statistical priors. The first group sets the range to $[\mu - 3\sigma, \mu + 3\sigma]$, the second group to $[\mu - 4\sigma, \mu + 4\sigma]$, the third group uses $[\min, \max]$, and the fourth group adopts $[\mu - \frac{1}{2}(\max - \min), \mu + \frac{1}{2}(\max - \min)]$. The results, presented in Tab.~\ref{tab:INIT}, show that the first group achieves the best performance, validating the effectiveness of our initialization strategy. To further verify this choice, we examine the distributions of learned $\hat{\alpha}$ and $\hat{\beta}$ after training (visualized in the \textit{Supplementary Material}) and observe that $\hat{\alpha}$ approximately follows $|\mu| + 3\sigma$ while $\hat{\beta}$ approximately equals $-\mu$. This convergence to our initialization prior confirms that DLS grounds quantization range construction in the statistical regularities of the data.

% \begin{table}
% % \vspace{2mm}
%   \centering
%   \begin{adjustbox}{width = \linewidth}
%   \begin{tabular}{@{}l c c >{\centering\arraybackslash}p{0.9cm} >{\centering\arraybackslash}p{0.9cm} >{\centering\arraybackslash}p{0.9cm} >{\centering\arraybackslash}p{1.3cm} >{\centering\arraybackslash}p{1.3cm}}
%     \toprule
%     % Method & Bit& \multicolumn{2}{c}{\textbf{Set5}}& \multicolumn{2}{c}{\textbf{Set14}}&\multicolumn{2}{c}{\textbf{B100}}&\multicolumn{2}{c}{\textbf{Urban100}}&\multicolumn{2}{c}{\textbf{Manga109}}\\
%     Method & $\alpha$&$\beta$& \textbf{Set5}& \textbf{Set14}&\textbf{B100}&\textbf{Urban100}&\textbf{Manga109}\\
%     % \cmidrule{3-7}
%     % &(w/a)&PSNR&SSIM&PSNR&SSIM&PSNR&SSIM&PSNR&SSIM&PSNR&SSIM\\
%     \midrule
%     % MambaIR-light \cite{guo2024mambair} & &&38.16&34&32.34&32.92&39.31\\
%     % \midrule
%     % Baseline & 4/4 &37.88&33.44&32.04&31.37&38.15\\
%     \rowcolor{gray!20}
%     GROUP1&$\alpha_1$&$\beta_1$&37.98&33.72&32.21&32.36&38.86\\  GROUP2&$\alpha_1$&$\beta_2$&37.03&32.65&31.46&29.66&36.50\\  GROUP3&$\alpha_2$&$\beta_1$&37.71&33.17&31.91&31.01&37.93\\ GROUP4&$\alpha_2$&$\beta_2$&37.68&33.17&31.90&30.98&37.92\\
%     \bottomrule
%   \end{tabular}
%   \end{adjustbox}
%   \caption{Ablation experiments for different designs of DLS.}
%   \label{tab:INIT}
% \end{table}

\subsection{Visualization}
\subsubsection{Visual Results.} In Fig.~\ref{fig:teaser}, we provide visual results of representative methods with scale $4\times$ on the Classic SR task in terms of 2-bit and 4-bit cases. We conduct comparisons against several advanced quantization methods, including LSQ \cite{esser2019learned}, QuantSR \cite{qin2023quantsr}, Quamba \cite{chiang2025quamba}, and MambaQuant \cite{xu2025mambaquant}. Our Q-MambaIR recovers more structural details and alleviates more blurring artifacts than other methods. Remarkably, the 4-bit quantized model achieves a visual quality close to that of the 32-bit full-precision MambaIR \cite{guo2024mambair}. These visual comparisons further demonstrate the effectiveness of Q-MambaIR, which is consistent with Tab.~\ref{tab:classic}.

\subsubsection{Outliers.} 
\begin{figure}
  \centering
    \includegraphics[width =\linewidth]{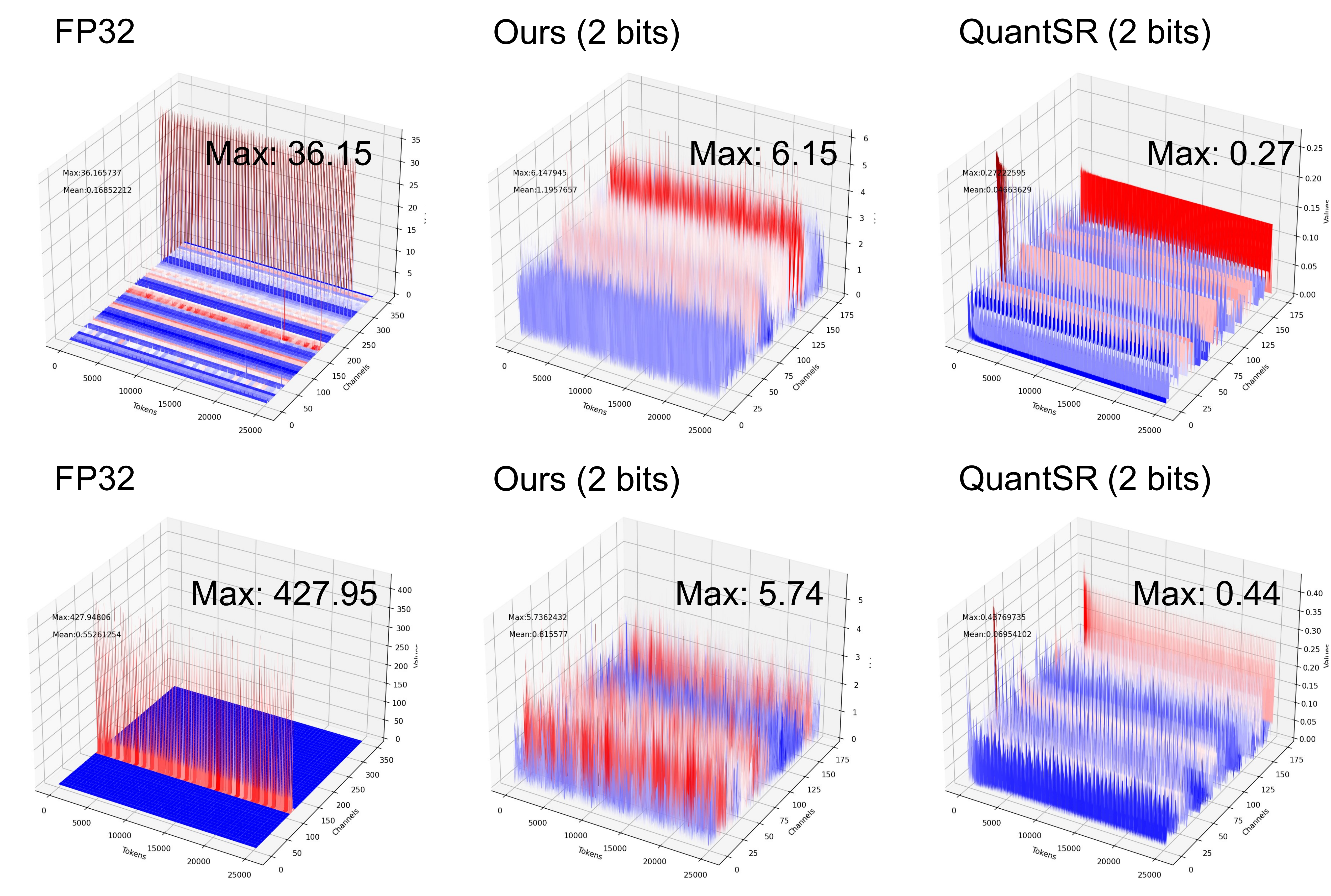}
    \caption{Visualization of activation distributions at the SSM output of \textit{Block0.Layer2} and \textit{Block0.Layer3} in 2-bit Q-MambaIR, on lightweight image SR ($\times$2). Our method preserves more extreme values than QuantSR~\cite{qin2023quantsr}.}
    \vspace{-3mm}
  \label{fig:out_y}
  \vspace{-2mm}
\end{figure}

In Fig.~\ref{fig:out_y}, using the same input, we visualize the output distributions of \textit{Block0.Layer2} and \textit{Block0.Layer3} in the SSM modules (ASSM) of the 2-bit Q-MambaIRv2~\cite{guo2025mambairv2}. Compared with QuantSR\cite{qin2023quantsr}, our method preserves a significantly larger number of extreme values within the SSM outputs. This suggests that our approach better balances detail retention with the preservation of high-magnitude responses, making it a more outlier-aware quantization strategy that is naturally suited for SSMs.

% \subsubsection{Learnable parameters.} \cref{fig:out_y} presents the statistical distribution of learned quantization parameters: (a) and (b) show the distributions of $\hat{\alpha}$ and $\hat\beta$ weights for trained DLS quantizers, while (c) displays the learned threshold ($\hat{T_i}$) distributions across RFA quantizers, revealing the learned parameter characteristics after training. From Figures (a) and (b), we observe that $\hat{\alpha}$ approximately follows $mean + 3·std$, while $\hat\beta$ approximately equals 
% $-mean$. This observation aligns with the first initialization group in \cref{tab:INIT}, and further validates the motivation behind DLS, linking quantization range construction to the data distribution, rather than simply learning the quantization bounds$\alpha$ and $\beta$ in an unconstrained manner. By grounding the learned parameters in statistical regularities, DLS avoids blindly fitting arbitrary ranges and instead forms quantization intervals that are both interpretable and distribution-aware.

\subsection{Limitation} Despite the significant performance gains brought about by Q-MambaIR under low-bit quantization, there remains a noticeable performance gap compared to its 32-bit full-precision counterpart. For example, fine-grained local textures around blurred edges in the SR task tend to suffer from visible artifacts or oversmoothing under ultra-low bitwidths settings. Bridging this gap remains a challenging and meaningful direction for future research.
\section{Conclusion}

In this paper, we propose Q-MambaIR, an accurate, efficient, and flexible Quantized Mamba for IR tasks, addressing critical challenges in low-bit quantization of SSM-based models. The proposed network employs a DLS to provide dynamic quantization ranges and an RFA to preserve fine-grained gradient information. Together, they effectively mitigate the issues of outlier accommodation, gradient mismatch, and parameter homogenization while preserving high-frequency textures. With extensive experiments, we demonstrate that Q-MambaIR outperforms existing state-of-the-art quantized IR networks in both accuracy and computational efficiency.

\section{Acknowledgement}
This work was supported by a grant from the Swiss National Supercomputing Centre (CSCS) under project ID lp12 on Alps. We acknowledge ISCRA for awarding this project access
to the LEONARDO supercomputer, owned by the EuroHPC
Joint Undertaking, hosted by CINECA (Italy).
\renewcommand{\maketitlesupplementary}{%
   \newpage
   \twocolumn[{%
       \centering
       \Large
       \textbf{\thetitle}\\[0.5em]
       Supplementary Material \\[1em]
       \large
       Yujie Chen\textsuperscript{1},
       Haotong Qin\textsuperscript{1}\dag,
       Zhang Zhang\textsuperscript{2},
       Michele Magno\textsuperscript{1},
       Luca Benini\textsuperscript{1},
        Yawei Li\textsuperscript{1,3}\dag\\[0.5em]
        \textsuperscript{1}ETH Z\"urich \quad
        \textsuperscript{2}Shenzhen Automotive Research Institute \quad \textsuperscript{3}Nanyang Technological University\\[0.3em]
        {\tt\small \{li.yawei.ai@gmail.com, haotong.qin@pbl.ee.ethz.ch\}}\\[1.5em]
   }]
}

\clearpage
\setcounter{page}{1}
\maketitlesupplementary

\begin{table*}
  \caption{Quantitative comparison on \textbf{lightweight image SR} with SOTA methods. }
  \label{tab:light}
% \vspace{-2mm}
  \centering
  \begin{adjustbox}{width=\textwidth}
  \begin{tabular}{@{}l c c c c c c c c c c c c }
    \toprule
    % \multirow{2}{*}{\textbf{Method}}
    % \multirow{2}{*}{\textbf{Method}} & \multirow{1}{*}{\textbf{Bit}}  & 
    % \multicolumn{2}{c}{\textbf{Set5}} & \multicolumn{2}{c}{\textbf{Set14}} & \multicolumn{2}{c}{\textbf{B100}} & \multicolumn{2}{c}{\textbf{Urban100}} & \multicolumn{2}{c}{\textbf{Manga109}}\\
    \multirow{2}{*}{\textbf{Method}} & \multirow{2}{*}{\textbf{Scale}} & Bit& \multicolumn{2}{c}{\textbf{Set5}}& \multicolumn{2}{c}{\textbf{Set14}}&\multicolumn{2}{c}{\textbf{B100}}&\multicolumn{2}{c}{\textbf{Urban100}}&\multicolumn{2}{c}{\textbf{Manga109}}\\
    \cmidrule{4-13}
    &&(w/a)&PSNR&SSIM&PSNR&SSIM&PSNR&SSIM&PSNR&SSIM&PSNR&SSIM\\
    \midrule
    \rowcolor{gray!20}
    MambaIR-light \cite{guo2024mambair} &$2\times$ &32/32 &38.16 &0.9610 &34.00 &0.9212 &32.34 &0.9017 &32.92 &0.9356 &39.31 &0.9779 \\
    % \hdashline
    MambaQuant* \cite{xu2025mambaquant} &$2\times$ &8/8 &38.05&0.9600&33.73&0.9174&32.28&0.9006&32.37&0.9301&38.72&0.9762\\
    Quamba* \cite{chiang2025quamba} &$2\times$ &8/8 &38.07&0.9600&33.81&0.92&32.28&0.9007&32.58&0.9322&38.92&0.9773\\
    % Quamba2* &$2\times$ &4/8 &37.70& 0.9579\\ 
    % Quamba2* &$2\times$ &4/6 &34.40& 0.9282\\ 
    % Quamba2* &$2\times$ &4/4 &24.69 & 0.5120\\ 
    \midrule
    % \hdashline
    MambaQuant* \cite{xu2025mambaquant}&$2\times$ &4/4 &25.62&0.5518&24.46&0.5276&24.72&0.5393&23.06&0.5292&24.62&0.5231\\
    Quamba* \cite{chiang2025quamba}&$2\times$ &4/4 &26.22&0.5778&25.02&0.5610&25.47&0.5833&23.53&0.5578&24.55&0.5285\\
    % MambaQuant &$2\times$ &4/4\\
    LSQ \cite{esser2019learned} &$2\times$ &4/4 &\textcolor{blue}{37.94} &\textcolor{blue}{0.9602} &\textcolor{blue}{33.71} &\textcolor{blue}{0.9189} &\textcolor{blue}{32.18} &\textcolor{blue}{0.8995} &\textcolor{blue}{32.28} &\textcolor{blue}{0.9304} &\textcolor{blue}{38.81} &\textcolor{blue}{0.9774} \\
    QuantSR \cite{qin2023quantsr} &$2\times$ &4/4 &37.88 &0.9598 &33.44 &0.9166 &32.04 &0.8977 &31.37 &0.9216 &38.15 &0.9756 \\
    Q-MambaIR &$2\times$ &4/4 &\textcolor{red}{37.99} &\textcolor{red}{0.9604} &\textcolor{red}{33.77} &\textcolor{red}{0.9193} &\textcolor{red}{32.22} &\textcolor{red}{0.9000} &\textcolor{red}{32.43} &\textcolor{red}{0.9314} &\textcolor{red}{38.91} &\textcolor{red}{0.9776} \\
    \midrule
    % Quamba &$2\times$ &2/2&----\\
    % MambaQuant &$2\times$ &2/2\\
    LSQ \cite{esser2019learned} &$2\times$ &2/2 &36.95 &0.9563 &32.63 &0.9088 &31.50 &0.8904 &29.87 &0.9033 &36.42 &0.9705 \\
    QuantSR \cite{qin2023quantsr} &$2\times$ &2/2 &\textcolor{blue}{37.31} &\textcolor{blue}{0.9576} &\textcolor{blue}{32.86} &\textcolor{blue}{0.9112} &\textcolor{blue}{31.67} &\textcolor{blue}{0.8928} &\textcolor{blue}{30.39} &\textcolor{blue}{0.9103} &\textcolor{blue}{37.09} &\textcolor{blue}{0.9728} \\
    Q-MambaIR &$2\times$ &2/2 &\textcolor{red}{37.39} &\textcolor{red}{0.9580} &\textcolor{red}{33.01} &\textcolor{red}{0.9129} &\textcolor{red}{31.80} &\textcolor{red}{0.8950} &\textcolor{red}{30.81} &\textcolor{red}{0.9161} &\textcolor{red}{37.43} &\textcolor{red}{0.9738} \\
    \midrule
    \midrule
    \rowcolor{gray!20}
    MambaIR-light \cite{guo2024mambair} &$3\times$ &32/32 &34.72 &0.9296 &30.63 &0.8475 &29.29 &0.8099 &29.00 &0.8689 &34.39 &0.9495\\
    % \hdashline
    Quamba* \cite{chiang2025quamba}&$3\times$ &8/8&35.54&0.9278&30.37&0.8427&28.15&0.8060&28.45&0.8572&33.67&0.9442\\
    MambaQuant* \cite{xu2025mambaquant}&$3\times$ &8/8&34.52&0.9274&30.36&0.8422&29.15 &0.8060&28.37&0.8556&33.58&0.9432\\
    % \hdashline
    \midrule
    Quamba* \cite{chiang2025quamba}&$3\times$ &4/4&23.09&0.4316&22.32&0.3944&22.13&0.3758&21.01&0.3987&22.41&0.4221\\
    MambaQuant* \cite{xu2025mambaquant}&$3\times$ &4/4&24.42&0.4878&23.26& 0.4423& 23.12&0.4216&21.45&0.4268&21.77&0.3952\\
    % MambaQuant &$3\times$ &4/4\\
    LSQ \cite{esser2019learned} &$3\times$ &4/4 &\textcolor{blue}{34.41} &\textcolor{blue}{0.9270} &\textcolor{blue}{30.34} &\textcolor{blue}{0.8424} &14.00 &0.3182 &\textcolor{blue}{28.23} &\textcolor{blue}{0.8551} &\textcolor{blue}{33.66} &\textcolor{blue}{0.9446} \\
    QuantSR \cite{qin2023quantsr} &$3\times$ &4/4 &34.34 &0.9265 &30.23 &0.8390 &\textcolor{blue}{29.02} &\textcolor{blue}{0.8023} &27.82 &0.8459 &33.03 &0.9407 \\
    Q-MambaIR &$3\times$ &4/4 &\textcolor{red}{34.46} &\textcolor{red}{0.9273} &\textcolor{red}{30.40} &\textcolor{red}{0.8434} &\textcolor{red}{29.15} &\textcolor{red}{0.8058} &\textcolor{red}{28.37} &\textcolor{red}{0.8576} &\textcolor{red}{33.78} &\textcolor{red}{0.9455} \\
    \midrule
    % Quamba &$3\times$ &2/2 &----\\
    % MambaQuant &$3\times$ &2/2 \\
    LSQ \cite{esser2019learned} &$3\times$ &2/2 &32.95 &0.9127 &29.39 &0.8240 &14.04 &0.3217 &26.28 &0.8058 &30.62 &0.9170 \\
    QuantSR \cite{qin2023quantsr} &$3\times$ &2/2 &\textcolor{blue}{33.34} &\textcolor{blue}{0.9170} &\textcolor{blue}{29.58} &\textcolor{blue}{0.8278} &\textcolor{blue}{28.60} &\textcolor{blue}{0.7922} &\textcolor{blue}{26.64} &\textcolor{blue}{0.8176} &\textcolor{blue}{31.36} &\textcolor{blue}{0.9259} \\
    Q-MambaIR &$3\times$ &2/2 &\textcolor{red}{33.58} &\textcolor{red}{0.9197} &\textcolor{red}{29.78} &\textcolor{red}{0.8318} &\textcolor{red}{28.73} &\textcolor{red}{0.7955} &\textcolor{red}{27.03} &\textcolor{red}{0.8285} &\textcolor{red}{31.98} &\textcolor{red}{0.9316} \\
    \midrule
    \midrule
    \rowcolor{gray!20}
    MambaIR-light \cite{guo2024mambair}&$4\times$ &32/32 &32.51 &0.8993 &28.85 &0.7876 &27.75 &0.7423 &26.75 &0.8051 &31.26 &0.9175\\
    % \hdashline
    Quamba* \cite{chiang2025quamba}&$4\times$ &8/8&32.30&0.8972&28.70&0.7834&27.67&0.7390&26.41&0.7944&30.74&0.9112\\
    MambaQuant* \cite{xu2025mambaquant}&$4\times$ &8/8&32.40 &0.8971&28.70&0.7835&27.67&0.7391&26.41&0.7942&30.73&0.9111\\
    \midrule
    % \hdashline
    Quamba* \cite{chiang2025quamba}&$4\times$ &4/4&22.47&0.4193&20.68&0.3298&20.33&0.2984&18.84&0.3200&21.23&0.3842\\
    MambaQuant* \cite{xu2025mambaquant}&$4\times$ &4/4 &19.72&0.3081&18.61&0.2445&19.00&0.2367&17.63&0.2508&17.52&0.2320\\
    % MambaQuant &$4\times$ &4/4\\
    LSQ \cite{esser2019learned} &$4\times$ &4/4 &\textcolor{blue}{32.16} &\textcolor{blue}{0.8935} &\textcolor{blue}{28.57} &\textcolor{blue}{0.7805} &\textcolor{blue}{27.57} &\textcolor{blue}{0.7358} &\textcolor{blue}{26.11} &\textcolor{blue}{0.7873} &\textcolor{blue}{30.37} &\textcolor{blue}{0.9063} \\
    QuantSR \cite{qin2023quantsr} &$4\times$ &4/4 &32.06 &0.8925 &28.37 &0.7757 &27.49 &0.7327 &25.55 &0.7674 &29.86 &0.8987 \\
    Q-MambaIR &$4\times$ &4/4 &\textcolor{red}{32.16} &\textcolor{red}{0.8941} &\textcolor{red}{28.62} &\textcolor{red}{0.7816} &\textcolor{red}{27.59} &\textcolor{red}{0.7367} &\textcolor{red}{26.17} &\textcolor{red}{0.7891} &\textcolor{red}{30.53} &\textcolor{red}{0.9080} \\
    \midrule
    % Quamba &$4\times$ &2/2 &l-ing\\
    % MambaQuant &$4\times$ &2/2\\
    LSQ \cite{esser2019learned} & $4\times$ & 2/2 & 30.84 & 0.8724 & 27.71 & 0.7597 & 27.03 & 0.7168 & 24.68 & 0.7345 & 27.86 & 0.8662 \\
    QuantSR \cite{qin2023quantsr} & $4\times$ & 2/2 & \textcolor{blue}{31.08} & \textcolor{blue}{0.8769} & \textcolor{blue}{27.84} & \textcolor{blue}{0.7633} & \textcolor{blue}{27.11} & \textcolor{blue}{0.7198} & \textcolor{blue}{24.82} & \textcolor{blue}{0.7419} & \textcolor{blue}{28.26} & \textcolor{blue}{0.8747} \\
    Q-MambaIR & $4\times$ & 2/2 & \textcolor{red}{31.15} & \textcolor{red}{0.8779} & \textcolor{red}{28.00} & \textcolor{red}{0.7665} & \textcolor{red}{27.18} & \textcolor{red}{0.7227} & \textcolor{red}{25.05} & \textcolor{red}{0.7499} & \textcolor{red}{28.67} & \textcolor{red}{0.8800} \\
%&31.1500&0.8779&28.4802&0.7665&27.1784&0.7226&25.0504&0.7499&28.6665&0.8800\\
    \bottomrule
  \end{tabular}
  \end{adjustbox}
\end{table*}

\begin{table}
\caption{Quantitative comparison on \textbf{SwinIR} and \textbf{EDSR} (4x).}
  \centering
  \begin{adjustbox}{width=.5\textwidth}
  \begin{tabular}{@{}l c c c c c}
    \toprule
   \textbf{Method} & Bit (w/a)& \textbf{Set5}& \textbf{Set14}&\textbf{B100}&\textbf{Urban100}\\
    % \cmidrule{4-13}
    % &&(w/a) \\ % &PSNR&SSIM&PSNR&SSIM&PSNR&SSIM&PSNR&SSIM&PSNR&SSIM \\
    \midrule
    \rowcolor{gray!20}
     EDSR \cite{esser2019learned}&32/32 & 32.46 & 28.72& 27.72 & 26.67\\
    % LSQ \cite{esser2019learned} &$4\times$ &4/4 &\textcolor{blue}{32.16} &\textcolor{blue}{0.8935} &\textcolor{blue}{28.57} &\textcolor{blue}{0.7805} &\textcolor{blue}{27.57} &\textcolor{blue}{0.7358} &\textcolor{blue}{26.11} &\textcolor{blue}{0.7873} &\textcolor{blue}{30.37} &\textcolor{blue}{0.9063} \\
    QuantSR \cite{qin2023quantsr}&4/4 &31.93 &28.37 &27.49 &25.55\\
    Ours &4/4 &\textcolor{red}{31.96}&\textcolor{red}{28.62}  &\textcolor{red}{27.59}&\textcolor{red}{26.17}  \\
    \midrule
    % Quamba &$4\times$ &2/2 &l-ing\\
    % MambaQuant &$4\times$ &2/2\\
    % LSQ \cite{esser2019learned} & $4\times$ & 2/2 & 30.84 & 0.8724 & 27.71 & 0.7597 & 27.03 & 0.7168 & 24.68 & 0.7345 & 27.86 & 0.8662 \\
    QuantSR \cite{qin2023quantsr} & 2/2 &31.51 & 27.84&27.11& 24.82\\
    Ours & 2/2 & \textcolor{red}{31.55} & \textcolor{red}{28.00}  & \textcolor{red}{27.18}& \textcolor{red}{25.05}  \\
    \midrule
    \midrule
    \rowcolor{gray!20}
    SwinIR \cite{liang2021swinir} &32/32 & 32.4 & 28.72 & 27.72  & 26.67\\
    % LSQ \cite{esser2019learned} &$4\times$ &4/4 &\textcolor{blue}{32.16} &\textcolor{blue}{0.8935} &\textcolor{blue}{28.57} &\textcolor{blue}{0.7805} &\textcolor{blue}{27.57} &\textcolor{blue}{0.7358} &\textcolor{blue}{26.11} &\textcolor{blue}{0.7873} &\textcolor{blue}{30.37} &\textcolor{blue}{0.9063} \\
    QuantSR \cite{qin2023quantsr}  &4/4 &32.11&28.37 &27.49 &25.55  \\
    Q-MambaIR &4/4 &\textcolor{red}{32.16} &\textcolor{red}{28.62} &\textcolor{red}{27.59} &\textcolor{red}{26.17}  \\
    \midrule
    % Quamba &$4\times$ &2/2 &l-ing\\
    % MambaQuant &$4\times$ &2/2\\
    LSQ \cite{esser2019learned} & 2/2 & 30.84& 27.71 & 27.03 & 24.68 \\
    QuantSR \cite{qin2023quantsr} & 2/2 & 31.08& 27.84& 27.11&24.82 \\
    Q-MambaIR & 2/2 & \textcolor{red}{30.87}& \textcolor{red}{28.00}& \textcolor{red}{27.18} & \textcolor{red}{25.05}  \\
%&31.1500&0.8779&28.4802&0.7665&27.1784&0.7226&25.0504&0.7499&28.6665&0.8800\\
    \bottomrule
  \end{tabular}
  \end{adjustbox}
  \label{tab:swinir}
\end{table}

\begin{table*}
\caption{Gaussian color image denoising}
\label{tab:DN}
\vspace{2mm}
  \centering
  % \setlength{\tabcolsep}{1pt}
  % \begin{adjustbox}{width=\textwidth}`  
  \begin{tabular}{@{} l c c c c c c c c c c}
    \toprule
    %     \toprule
    % \multirow{2}{*}{\textbf{Method}}
    % \multirow{2}{*}{\textbf{Method}} & \multirow{1}{*}{\textbf{Bit}}  & 
    % \multicolumn{2}{c}{\textbf{Set5}} & \multicolumn{2}{c}{\textbf{Set14}} & \multicolumn{2}{c}{\textbf{B100}} & \multicolumn{2}{c}{\textbf{Urban100}} & \multicolumn{2}{c}{\textbf{Manga109}}\\
    Method &Bit& \multicolumn{2}{c}{\textbf{BSD68}}& \multicolumn{2}{c}{\textbf{Kodak24}}&\multicolumn{2}{c}{\textbf{McMaster}}&\multicolumn{2}{c}{\textbf{Urban100}}\\
    \cmidrule{3-10}
    &(w/a)&$\sigma$=15&$\sigma$=25&$\sigma$=15&$\sigma$=25&$\sigma$=15&$\sigma$=25&$\sigma$=15&$\sigma$=25\\
    % Method &Bit (w/a)&{\textbf{BSD68}}&{\textbf{Kodak24}}&{\textbf{McMaster}}&{\textbf{Urban100}}\\
    \midrule
    \rowcolor{gray!20}
    MambaIR \cite{guo2024mambair}  &32/32 &34.48&31.80 &35.42&32.91&35.70&33.35&35.37&32.99\\
    \midrule
    Quamba* \cite{chiang2025quamba}  &8/8 &30.10&30.11&31.29&31.30&31.59&31.60&30.34&30.34\\
    MambaQuant* \cite{xu2025mambaquant}  &8/8&30.49&26.73&30.79&26.31&29.48&27.09&28.62&25.50\\
    % \hdashline
    \midrule
    Quamba* \cite{chiang2025quamba} &4/4 &28.26&24.97&28.60&25.29&28.53&24.71&27.58&24.11\\
    MambaQuant* \cite{xu2025mambaquant} &4/4&20.49&21.00&20.17&20.16&20.43&20.27&22.98&21.10\\
    LSQ \cite{esser2019learned}&4/4 &34.27&29.93 &\textcolor{blue}{35.14}&30.48 &\textcolor{blue}{35.32}&30.06 &\textcolor{blue}{34.78}&28.90\\
    QuantSR \cite{qin2023quantsr} &4/4 &\textcolor{blue}{34.29} &\textcolor{blue}{31.59} &34.97&\textcolor{blue}{32.59} &34.38&\textcolor{blue}{32.91} &34.07&\textcolor{blue}{32.33} \\
    Q-MambaIR&4/4 &\textcolor{red}{34.36}&\textcolor{red}{31.73} &\textcolor{red}{35.26}&\textcolor{red}{32.81} &\textcolor{red}{35.49}&\textcolor{red}{33.21} &\textcolor{red}{35.03}&\textcolor{red}{32.81} \\
    \midrule
    LSQ \cite{esser2019learned} &2/2 &32.68&24.85 &33.07&25.09 & 32.37&25.34 &31.66&24.21\\
    QuantSR \cite{qin2023quantsr} &2/2 &\textcolor{blue}{33.76}&\textcolor{blue}{31.04} &\textcolor{blue}{34.43}&\textcolor{blue}{31.87} &\textcolor{blue}{34.40}&\textcolor{blue}{31.96} &\textcolor{blue}{33.63}&\textcolor{blue}{30.99} \\
    Q-MambaIR  &2/2 &\textcolor{red}{33.81}&\textcolor{red}{31.09} &\textcolor{red}{34.51}&\textcolor{red}{31.95} &\textcolor{red}{34.49}&\textcolor{red}{32.06} &\textcolor{red}{33.69}&\textcolor{red}{31.11} \\
    \bottomrule
  \end{tabular}
  % \end{adjustbox}
\end{table*}

\section{Technical Appendices and Supplementary Material}
\label{sec:Appendix}
We provide additional information in this supplementary material. In \cref{app:exp}, we provide detailed experimental details to aid in the reproducibility of our method.
In \cref{app:re}, we provide more experimental results to aid in comparison and further validate the validity of our approach.
% Technical appendices with additional results, figures, graphs and proofs may be submitted with the paper submission before the full submission deadline (see above), or as a separate PDF in the ZIP file below before the supplementary material deadline. There is no page limit for the technical appendices.

\subsection{Experimental Setup and Training Details}
\label{app:train}
\label{app:exp}
We adopt a task-specific training strategy across various Image Restoration tasks. Below, we detail the training configurations for each scenario.

\subsubsection{Classic Image Super-Resolution (SR)}
For classic SR, we adopt the following training configuration: batch size 4, learning rate $1 \times 10^{-4}$, $L_1$ loss, GT size $192 \times 192$, and 6 Residual State Space Blocks (ss2D + CAB). We employ DLS to dynamically adjust the quantization range and mitigate peak truncation loss caused by outliers, while $L_1$ loss helps retain fine image details during optimization.

\subsubsection{Lightweight Image Super-Resolution}
For lightweight SR, we use a smaller configuration suited for edge deployment: batch size 2, learning rate $2 \times 10^{-4}$, $L_1$ loss, GT size $192 \times 192$, and 4 Residual State Space Blocks. RFA enables flexible rounding via adaptive thresholds, preserving high-frequency details under reduced model depth.

\subsubsection{Image Denoising (DN)}
For image denoising, we use batch size 4, learning rate $1 \times 10^{-4}$, Charbonnier loss, GT size $128 \times 128$, and 6 Residual State Space Blocks. We combine DLS and RFA to mitigate information loss under aggressive quantization while preserving texture and structure.

\subsubsection{JPEG Compression Artifact Reduction (JPEG CAR)}
For JPEG CAR, we use batch size 4, learning rate $1 \times 10^{-4}$, Charbonnier loss, GT size $128 \times 128$, and 6 Residual State Space Blocks. RFA-driven quantization helps preserve edge structures and fine details that are typically degraded by rigid quantization schemes.

\subsubsection{Training Milestones}
The learning rate is decayed at iterations 10{,}000, 15{,}000, 17{,}500, and 18{,}750 to refine optimization throughout training.

\subsection{Experimental Results}
\label{app:tasks}
\label{app:re}
In this section, we provide additional experiments to further demonstrate the effectiveness and general applicability of Q-MambaIR in a wide range of IR tasks. \\
\subsubsection{Comparison on extended tasks} 
\noindent \textbf{Comparison on Lightweight Image SR.}
We apply our method to quantize the MambaIR-light models. In the 4-bit case, our Q-MambaIR outperforms LSQ \cite{esser2019learned} by up to 0.21\,dB PSNR on the $2\times$ scale Urban100 dataset. In the 2-bit case, our Q-MambaIR outperforms LSQ \cite{esser2019learned} by up to 2.49\,dB PSNR on the $2\times$ scale Urban100 dataset. Our method provides a possible solution for ultra-low bit quantization. As shown in \cref{tab:light}, this addition further validates the scalability and adaptability of our method across different levels of upsampling.

\noindent \textbf{Comparison on Gaussian Denoising.} We extend our gaussian denoising experiment with two other noise level settings by 15 and 25 in Tab.~\ref{tab:DN}. It can be seen that our Q-MambaIR significantly outperforms the state-of-the-art methods at different noise levels, demonstrating the effectiveness and robustness of our method.

\subsubsection{Comparison on other IR architectures.}
\label{app:arch}
We further quantize both CNN-based and Transformer-based IR architectures with our method. As shown in \cref{tab:swinir}, our model also outperforms other quantization methods across different architectures in both 2-bit and 4-bit settings. The best performances are highlighted in red.

\section{Quantization Details}
\subsection{Quantized Operators}

\subsubsection{Weight Quantization} 
We employ a learnable uniform quantization RFA quantizer for all weight parameters. The complete pseudocode of the RFA quantizer, which follows the forward-discretization and backward-approximation strategies described in Sec.~2.3 of the main paper, is provided in Algorithm~\ref{alg:rfa}.
\begin{algorithm}[ht]
    \caption{Range-Floating Flexible Allocator (RFA)}
    \label{alg:rfa}
    \textbf{Input}: Input tensor $x$, learnable thresholds $T = \{T_0, T_1, ..., T_N\}$, quant levels $Q = \{q_0, q_1, ..., q_{N-1}\}$\\
    \textbf{Output}: Quantized output $\hat{x}$
    \begin{algorithmic}[1]
        \STATE Initialize $x_\text{forward} \gets x$, $x_\text{backward} \gets x$
        \FOR{$i = 0$ to $N-1$}
            \IF{$i = 0$}
                \STATE Set forward threshold $\tau_f \gets T_0$, backward threshold $\tau_b \gets T_0$
                \STATE Set quantization step $\Delta q \gets q_1 - q_0$
                \STATE Compute interpolated gradient path: $y \gets \frac{\Delta q}{T_1 - T_0} x + q_0 - \frac{\Delta q}{T_1 - T_0} T_0$
                \STATE Update $x_\text{forward} \gets \text{where}(x > \tau_f, q_0, q_0)$
                \STATE Update $x_\text{backward} \gets \text{where}(x > \tau_b, y, 0.1x + q_0 - 0.1 T_0)$
            \ELSE
                \STATE Set forward threshold $\tau_f \gets \frac{T_{i-1} + T_i}{2}$, backward threshold $\tau_b \gets T_i$
                \IF{$i < N-1$}
                    \STATE $\Delta q \gets q_{i+1} - q_i$
                    \STATE $y \gets \frac{\Delta q}{T_{i+1} - T_i} x + q_i - \frac{\Delta q}{T_{i+1} - T_i} T_i$
                \ELSE
                    \STATE $y \gets x$ \COMMENT{last interval, identity fallback}
                \ENDIF
                \STATE Update $x_\text{forward} \gets \text{where}(x > \tau_f, q_i, x_\text{forward})$
                \STATE Update $x_\text{backward} \gets \text{where}(x > \tau_b, y, x_\text{backward})$
            \ENDIF
        \ENDFOR
        \STATE Final slope fallback: $x_\text{backward} \gets \text{where}(x > T_{N-1}, 0.1x + q_{N-1} - 0.1 T_{N-1}, x_\text{backward})$
        \STATE Combine forward and backward paths:
        \[
        \hat{x} \gets \text{detach}(x_\text{forward}) + x_\text{backward} - \text{detach}(x_\text{backward})
        \]
        \STATE \textbf{return} $\hat{x}$
    \end{algorithmic}
\end{algorithm}

The \textbf{soft forward threshold} is implemented as the variable $T$, representing the set of adaptive boundaries $\{T_0, T_1, \dots, T_N\}$ that define the partitioning of the input space. These thresholds are trained jointly with the model and allow the quantization scheme to dynamically adjust to the underlying data distribution.

% The \textbf{quantization centers} are predefined integer values stored in the list \texttt{additive\_pot = [-2^{n-1}, \dots, 2^{n-1}-1]}, where $n$ denotes the quantization bit-width. These represent the fixed output levels $q_i$ to which each input value may be mapped.

The forward hard rounding responsible for discretizing the input values, is realized via the operation $x_{forward} = torch.where(x > threshold, q_{new}, q_{old})$, which assigns to each input the nearest quantization level $q_i$ in a uniform distribution. For an $n$-bit quantizer, we define the set of quantization centers as a symmetric integer range:
\[
\mathcal{Q} = \left\{q_i \,\middle|\, q_i \in \mathbb{Z},\ -2^{n-1} \leq q_i < 2^{n-1} \right\}.
\]
 This emulates the behavior of non-uniform to uniform rounding.

To enable efficient gradient flow during quantization-aware training, we adopt a two-phase gradient approximation strategy. 
The \textbf{soft backward approximation} is implemented using a piecewise linear interpolation. Within the active transition interval $[T_0, T_1]$, we apply a linear interpolation between $T_0$ and $T_1$:
\[
y = \frac{\Delta q}{T_1 - T_0} x + q_0 - \frac{\Delta q}{T_1 - T_0} T_0
\]
where $\Delta q = q_1 - q_0$ is the quantization level spacing.This design ensures that inputs near the quantization boundaries receive meaningful gradients during training, thereby mitigating the discontinuities introduced by hard rounding.

To address the issue of vanishing gradients in flat quantization regions, we introduce a flat region fallback mechanism, which assigns a small constant slope (0.1) to regions where the quantization function would otherwise be flat and nondifferentiable. 
Outside this interval, to avoid vanishing gradients, a flat-slope fallback is used:
\[
x_\text{backward} = \begin{cases}
y & \text{if } x > \tau_b \\
0.1x + q_0 - 0.1 T_0 & \text{otherwise}
\end{cases}
\]

Finally, we decouple the forward and backward behaviors of the quantization function. This is expressed as $x_{forward}.detach() + x_{backward} - x_{backward}.detach()$, which allows the forward pass to use discrete quantized values while the backward pass flows through the soft approximation. This technique is critical for enabling effective end-to-end optimization. We also implemented a CUDA acceleration for this part to release the training burden.

Overall, this code-level design enables RFA to perform discretization-aware learning while preserving gradient flow, making it particularly suitable for state space models and image restoration tasks that are sensitive to quantization artifacts.

\subsubsection{Activation Quantization} We employ DDA quantizer for all activations in both convolutional and linear layers. Especially for SSM components, we integrate this adaptive mechanism into all inputs of selective scan and 
all main operator in VSSM module. It can dynamically adjust quantized parameters based on input distribution features and handle dynamic range variations in SSM computations:
\begin{align}
\alpha &= |w_1 \cdot \mu + w_2 \cdot \sigma + w_3 \cdot x_{\min} + w_4 \cdot x_{\max}| \\
\beta &= w_{21} \cdot \mu + w_{22} \cdot \sigma + w_{23} \cdot x_{\min} + w_{24} \cdot x_{\max}
\end{align}
where $\mu$ and $\sigma$ are the mean and standard deviation, and $w_i$ are learnable parameters.

% \subsection{Detailed quantized architectures}
% \section{Quantization Strategy}

% We present a comprehensive quantization framework for the MambaIR architecture that achieves significant model compression while maintaining reconstruction quality. Our approach employs a hybrid quantization strategy combining uniform quantization for standard operations with DDA quantization for SSM components.

\subsection{Quantized Modules}

Our framework quantizes all major components: (1) all convolutional layers in CAB and residual connections, (2) linear projections in SS2D modules, (3) all SSM internal states and parameters (input states, time steps, state matrices, skip connections), and (4) projection matrices for SSM parameters. We demonstrate effective quantization down to 2-bit precision, achieving up to 16× model compression while maintaining competitive performance.

\section{Jetson Orin Deployment Workflow}
\label{sec:jetson_deployment}

This section provides a complete overview of our deployment workflow on the NVIDIA Jetson Orin platform. The pipeline is designed to ensure stable inference under constrained compute resources, and consists of three major stages: ONNX export, TensorRT engine construction, and on-device inference. We additionally describe practical considerations that arise when deploying low-bit models under real-world embedded hardware constraints.

\subsection{Prerequisites}

The deployment environment relies on the software and hardware components listed below:

\begin{itemize}
    \item \textbf{NVIDIA Jetson Orin device} with JetPack SDK (including CUDA, cuDNN, TensorRT, and system libraries).
    \item \textbf{Docker environment} providing a reproducible runtime with CUDA, PyTorch, and TensorRT support. We adopt a custom image (\texttt{realdn:with\_tensorrt}) to ensure consistent operator availability.
    \item \textbf{PyTorch model checkpoint} (\texttt{.pth}) containing trained FP32 or quantized model weights.
    \item \textbf{TensorRT toolkit} and the command-line utility \texttt{trtexec} for building optimized inference engines.
\end{itemize}

This setup ensures that the model can be exported, converted, and executed consistently across different Jetson Orin units.

\subsection{Step 1: ONNX Model Export}

The deployment pipeline begins with exporting the PyTorch model into the ONNX format. This step is executed either on the host machine or within the Docker container. During export, we:

\begin{itemize}
    \item load the PyTorch checkpoint and perform module re-wrapping if required by TensorRT;
    \item trace or script the model to handle dynamic control flow;
    \item export the model using \texttt{torch.onnx.export} with ONNX opset 18 to match TensorRT's supported operator schema;
    \item validate the exported ONNX graph using \texttt{onnxsim} or \texttt{onnxruntime} to ensure graph correctness.
\end{itemize}

A valid ONNX file is required for the following optimization step.

\subsection{Step 2: TensorRT Engine Building}

Given the exported ONNX model, we build a TensorRT engine optimized for Jetson Orin’s GPU architecture. TensorRT parses the ONNX graph, selects the best available kernels, performs layer fusion, and generates a platform-specific inference engine.

Engine construction is performed using the \texttt{trtexec} utility with the following key configuration options:

\begin{itemize}
    \item \texttt{--optShapes=input:1x3x256x256} to specify the optimal static input resolution;
    \item \texttt{--memPoolSize=workspace:4096M} (FP16) or \texttt{8192M} (INT8) to allocate sufficient workspace for kernel fusion;
    \item \texttt{--builderOptimizationLevel=1} to ensure broad operator compatibility during graph fusion;
    \item \texttt{--int8} to enable low-precision execution.
\end{itemize}

TensorRT produces a serialized engine (\texttt{.plan}) file, which is directly executable on the device.

Overall, this deployment pipeline ensures that both FP16 and INT8 variants of our model can be executed reliably on Jetson Orin while achieving the performance reported in the main paper.

\subsection{Explanation and Summary}

We report the end-to-end latency of full-precision MambaIR and its int4 quantized variants on the Jetson Orin platform. We tested them with our \textbf{INT8 deployment pipeline}. Below, we provide a concise explanation of why all 4-bit models exhibit similar latency and why Q-MambaIR introduces no overhead.

Although our models are nominally quantized to 4-bit weights and activations, the current \textit{TensorRT} pipeline on Jetson Orin only supports INT8 deployment. Consequently, both the FP32 MambaIR model and the 4-bit variants are internally executed using \textbf{INT8/FP16 mixed-precision kernels}. 

Despite this, the 4-bit models still exhibit lower latency (110--121\,ms) compared to the original FP32 model (198.21\,ms). This speedup arises not from true 4-bit arithmetic, but from operator-level optimizations introduced in the Q-MambaIR module, including:

\begin{itemize}
    \item \textbf{Pre-quantized weight layout}: For RFA-based weights, the quantization is applied offline, allowing the engine to directly consume quantized weights without additional preprocessing. This reduces memory accesses and kernel overhead.
    \item \textbf{Activation computation optimizations}: For DDA, the input activations are pre-processed to capture their statistical characteristics, and the selective-scan operations are reorganized to improve memory coalescing and parallel execution on the GPU.
\end{itemize}

As a result, although the actual computation still uses INT8 kernels, the optimized operator structure allows the nominal 4-bit models to achieve lower latency than the unoptimized FP32 model. Differences among Baseline-4bit, DLS-4bit, RFA-4bit, and Q-MambaIR are negligible in runtime, explaining the similar measured latencies across these variants.

\textbf{Operator Rearrangement for Efficient Deployment}

To enable efficient deployment of our 4-bit models on Jetson Orin, we introduce two implementation optimizations within Q-MambaIR. First, for the weights involved in the RFA-based rounding scheme, we apply the 4-bit transformation \emph{offline} and directly deploy the quantized weights to the edge device. This eliminates any on-device quantization overhead and ensures that the selective-scan operator consumes already–quantized parameters.

Second, for the activation-side transformation in DDA, we optimize the computation of input activations following techniques similar to those used in \cite{gu2019fast}. Specifically, we restructure the activation computation to reduce redundant operations while accurately capturing the statistical characteristics of the input tensors. This improves numerical fidelity under low-bit quantization without introducing additional kernels or memory-transfer cost.

Together, these optimizations allow Q-MambaIR to retain the same latency as the baseline 4-bit model while benefiting from more stable quantization behavior and improved restoration accuracy. 
{
    \small
    \bibliographystyle{ieeenat_fullname}
    \bibliography{main}
}
\end{document}